\documentclass{article}

\usepackage{arxiv}
\usepackage{authblk}

\usepackage[utf8]{inputenc} 
\usepackage[T1]{fontenc}    
\usepackage{hyperref}      
\usepackage{url}          
\usepackage{booktabs}      
\usepackage{amsfonts}      
\usepackage{nicefrac}       
\usepackage{microtype}      
\usepackage{lipsum}		
\usepackage{graphicx}
\usepackage[round,authoryear]{natbib}
\usepackage{doi}
\usepackage[acronym, nogroupskip, nopostdot, nomain]{glossaries}
\makeglossaries
\usepackage{makecell}
\usepackage{pifont}
\usepackage{siunitx}

\usepackage{multirow}
\usepackage{tabularx}
\usepackage{float}
\usepackage{threeparttable}
\usepackage{setspace}
\usepackage{array} 

\newcolumntype{Y}{>{\hspace{0pt}\centering\arraybackslash}X}
\newcolumntype{C}[1]{>{\centering\arraybackslash}p{#1}}

\newacronym{viirs}{VIIRS}{Visible Infrared Imaging Radiometer Suite}
\newacronym{modis}{MODIS}{Moderate Resolution Imaging Spectroradiometer}
\newacronym{nasa}{NASA}{National Aeronautics and Space Administration}
\newacronym{noaa}{NOAA}{National Oceanic and Atmospheric Administration}
\newacronym{goes}{GOES}{Geostationary Operational Environmental Satellite}
\newacronym{ml}{ML}{machine learning}
\newacronym{us}{US}{United States}
\newacronym{conus}{CONUS}{contiguous United States}
\newacronym{mtbs}{MTBS}{Monitoring Trends in Burn Severity}
\newacronym{feds}{FEDS}{Fire Event Data Suite}
\newacronym{frp}{FRP}{fire radiative power}
\newacronym{rtma}{RTMA}{Real-Time Mesoscale Analysis}
\newacronym{hrrr}{HRRR}{High Resolution Rapid Refresh}
\newacronym{nam}{NAM}{North American Mesoscale}
\newacronym{nfdrs}{NFDRS}{National Fire Danger Rating System}
\newacronym{gsi}{GSI}{Gridpoint Statistical Interpolation}
\newacronym{crs}{CRS}{coordinate reference system}
\newacronym{npp}{NPP}{National Polar-orbiting Partnership}
\newacronym{sedd}{SEDD}{Score Entropy Discrete Diffusion}
\newacronym{era5}{ERA5}{ECMWF Reanalysis v5}
\newacronym{ecmwf}{ECMWF}{European Centre for Medium-Range Weather Forecasts}
\newacronym{farsite}{FARSITE}{Fire Area Simulator}
\newacronym{elmfire}{ELMFIRE}{Eulerian Level Set Model of FIRE spread}

\title{PyroStack: A Multi-Band Spatio-Temporal Sub-Daily Dataset for Wildfires in the United States}

\date{} 

\author[1\thanks{Correspondence to akondur@uci.edu}]{Arya Kondur}
\author[2]{Giosue Migliorini}
\author[3]{Cameron Schmitt}
\author[1,4]{Francesco Immorlano}
\author[1]{Tairan Wang}
\author[5]{Rebecca C. Scholten}
\author[3,5]{Efi Foufoula-Georgiou}
\author[6]{Gary Johnson}
\author[7]{Chris Lautenberger}
\author[6]{Valentin Waeselynck}
\author[6]{J. Shane Romsos}
\author[7]{Kasra Shamsaei}
\author[5,8,9]{Alejandro Tejedor}
\author[3]{Tianjia Liu}
\author[5]{Yang Chen}
\author[1,2]{Padhraic Smyth}
\author[3,5]{James T. Randerson}

\affil[1]{Department of Computer Science, University of California, Irvine, CA 92697, USA}
\affil[2]{Department of Statistics, University of California, Irvine, CA 92697, USA}
\affil[3]{Department of Civil and Environmental Engineering, University of California, Irvine, CA 92697, USA}
\affil[4]{Learning the Earth with Artificial Intelligence and Physics (LEAP) Center, Columbia University, New York, NY 10027, USA}
\affil[5]{Department of Earth System Science, University of California, Irvine, CA 92697, USA}
\affil[6]{Spatial Informatics Group, Pleasanton, CA 94566, USA}
\affil[7]{CloudFire Inc., Auburn, CA 95603, USA}
\affil[8]{Institute for Biocomputation and Physics of Complex Systems (BIFI), University of Zaragoza, Spain}
\affil[9]{Department of Theoretical Physics, University of Zaragoza, Spain}

\begin{document}
\maketitle
\begin{abstract}
Wildfires are an increasing hazard to ecosystems, air quality, and human systems, creating a growing need for datasets that support systematic development and evaluation of models for predicting fire spread across diverse landscapes. Effective prediction requires integrating meteorological conditions, fuels, vegetation, and topography at spatial and temporal resolutions suitable for both physical simulation and data-driven approaches. However, existing datasets often lack the resolution and coverage needed to capture these interacting controls. The PyroStack dataset addresses this gap by providing a harmonized, event-based collection of wildfire and environmental data across the contiguous United States and Alaska. It integrates satellite-derived fire observations with atmospheric reanalysis, vegetation, fuel characteristics, and topographic information into a unified framework spanning 6994 wildfires that occurred between 2012 and 2024 across a wide range of ecosystems and climate conditions. PyroStack offers spatial resolutions ranging from 30 m to 9 km and hourly temporal resolution, along with fire progression data at 12-hour intervals to support model initialization and evaluation. By combining broad spatial coverage with fine spatial and temporal detail, the dataset enables systematic analysis of wildfire dynamics and supports both physics-based and machine learning approaches, providing a foundation for benchmarking and improving fire spread models, with future extensions aimed at incorporating additional regions and fire suppression data streams to further advance wildfire prediction.
\end{abstract}

\section{Introduction}

Improving our ability to predict wildfire spread is an important endeavor, given that the impacts and severity of wildfires are increasing globally \citep{TeymoorSeydi_2025_human_exposure, xu_2023_exposure, jones_global_2024}. Wildfires threaten vulnerable ecosystems, contribute to global warming, deteriorate air quality, and pose risks to human lives and property. For example, their direct and indirect costs have been estimated at over $100$ billion USD in a single year in California alone \citep{wang_2021_economic}, and wildfire effects on air quality may be responsible for hundreds of thousands of deaths each year globally \citep{johnston_2012_mortality, xu_2023_exposure}. Given future climate change  projections and the expansion of the wildland-urban interface \citep{schug_global_2023, chen_wildfire_2024}, the impacts of wildfires on humans and ecosystems are likely to continue to escalate \citep{Gallo2025climatechange,Yang2026biodiversity}.

Anticipating how fires behave and spread, at different spatial and temporal scales, is critical for both suppression and land management. Improvements in models and forecasts are needed not only to support real-time decision-making during large wildfires, but also to guide the design and placement of fuel treatments. More fundamentally, understanding how fuel composition, meteorology, and fire dynamics interact is essential for characterizing the complex feedbacks that can occur between wildfires and climate change \citep{maciasfauria_2010_climate}. Physics-based fire behavior models, which often include empirical parameterizations that describe the fire spread rate as a function of fuel composition, moisture content, topography, and wind velocity, have traditionally been used to predict the rate and direction of fire spread at landscape or regional scales \citep{finney_1998_farsite, finney_2011_ensemble, lautenberger_2013_ELMFIRE, kelso_2015_aus_simulator}. However, these models are often limited by scarce high-resolution data, high computational costs, and challenges in systematic validation. At the same time, advances in \gls{ml} provide an opportunity to gain insight into wildfire processes and to improve forecasts of fire activity \citep{jain_2020_ml_review}. However, the prediction accuracy of such data-driven \gls{ml} approaches remains fundamentally constrained by the availability and quality of the training data.

In recent years, an increasing number of Earth system data streams have become available and have the potential to improve our ability to predict the progression of wildfires. Remote sensing observations from low-earth orbit satellites, such as those gathered from the \gls{modis} sensor on Aqua and Terra satellites and the \gls{viirs} sensor \citep{barnes_1998_modis, schroeder_2014_viirs} on \gls{noaa} satellites, provide twice-daily global observation of wildfire activity. The resulting datasets have enabled improved tracking of wildfires and smoke, as well as better quantification of changing fire patterns and their impacts on climate and air quality \citep{artes_2019_dataset, andela_2019_atlas, balch_2020_fired, chen_2022_viirs, Scholten_2024_environment}. A second major source of remote sensing data originates from geostationary satellites. Across the \gls{us}, the \gls{goes} system \citep{schmit_2017_goesr}, consisting of multiple geosynchronous equatorial satellites, provides a wealth of continuous near real-time data used for fire and smoke monitoring and forecasting \citep{liu_2024_progression, rolph_2009_rave}. Complementary geospatial and meteorological data sources provide essential information on the environmental conditions that govern fire behavior \citep{Li_2022_goes_frp}. For example, \gls{era5}, a global atmospheric reanalysis produced by the \gls{ecmwf}, provides hourly estimates of meteorological variables such as temperature, humidity, precipitation, and wind \citep{hersbach_2020_era5} and fire weather indices \citep{Vitolo_2020_era5}. Reanalysis products combine model simulations with assimilated observations to generate spatially and temporally complete records of the atmospheric state, making them particularly valuable for characterizing the weather conditions associated with the fire spread. At a finer spatial resolution, the LANDFIRE program provides a geospatial database of vegetation type, wildland fuel characteristics, and topography across the \gls{us} at \SI{30}{\meter} resolution \cite{reeves_2009_landfire}. These data products serve as essential inputs for operational fire-spread models in the United States, providing canopy height and structural information for live and dead fuel layers, which are necessary to understand fire-environment interactions. 

Translating these increasingly rich and heterogeneous data streams into advances in fire modeling requires harmonizing them into a consistent, analysis-ready format that supports both physics-based and data-driven approaches. This harmonization step remains a major bottleneck for wildfire researchers looking to improve predictions of fire spread, as well as assessments of wildfire risk and hazard \citep{jain_2020_ml_review, kondylatos_2023_mesogeos, oliveira_2023_near}. In response, multiple research groups have developed integrated wildfire datasets designed to address this challenge (e.g., \cite{lahrichi_2025_wsts+, zhao_2025_tssatfire, muller_high-resolution_2026}). These products differ substantially in their spatial domain, spatial and temporal resolution, and the number and type of variables included in the data stack (see Table \ref{table:datasetComparison}). Many of the existing datasets provide meteorological variables at daily resolution, which is insufficient for driving state-of-the-art physical fire spread models such as \gls{elmfire} \citep{lautenberger_2013_ELMFIRE}. These models require high-frequency (i.e., $\sim$hourly) inputs, particularly for wind and fuel moisture. In addition, accurately representing topographic controls and fuel heterogeneity requires land surface data with relatively fine spatial resolution ($\sim$30 m). However, many current datasets offer only coarser resolutions or lack such data entirely.

To address these limitations, we introduce PyroStack, a standardized, multivariate, event-based spatiotemporal dataset. PyroStack provides nested spatial resolutions from 30 m to 9 km and temporal resolutions up to hourly, integrating topography, meteorology, vegetation, and fuel characteristics with co-located fire observations at 12-hour intervals. The dataset is specifically designed to support both (a) the initialization, validation, and benchmarking of physics-based fire spread models, and (b) the development of \gls{ml} approaches that operate at comparable spatial and temporal scales. It includes nearly 7,000 fires across the contiguous United States and Alaska, spanning diverse ecosystems and climatic regimes. To demonstrate its utility, we present below example simulations of a representative wildfire using both physics-based and machine learning models. A central objective of PyroStack is to provide a common testbed for systematic comparison and benchmarking of physics-based and \gls{ml} modeling approaches. PyroStack improves over existing wildfire datasets by providing a substantially larger number of fire events, broader spatial coverage across the contiguous United States and Alaska, finer spatial resolution for key fuel and topographic layers (30 m), finer temporal resolution, including hourly meteorological forcing, and the inclusion of fire area and active fireline observations at 12-hour intervals that enable multiple fire spread model initializations for a single wildfire. 

\begin{table}[H]
    \centering
    \scriptsize
    \begin{threeparttable}
    \setlength{\tabcolsep}{2.5pt}
    \renewcommand{\arraystretch}{1.15}
    \caption{Comparison of existing fire tracking datasets with the PyroStack dataset introduced in this study.}
    \label{table:datasetComparison}
    \begin{tabularx}{\linewidth}{
        Y          
        Y          
        Y          
        Y         
        C{0.95cm}  
        C{0.85cm}  
        C{0.65cm}  
        Y          
        C{0.85cm}  
        C{0.95cm}  
    }
        \toprule
        \textbf{Name} &
        \textbf{Reference} &
        \textbf{Region} &
        \textbf{Temporal res.} &
        \textbf{Spatial res.} &
        \textbf{Num. of fire trajectories}\tnote{1} &
        \textbf{Num. of vars} &
        \textbf{Variable types}\tnote{2} &
        \textbf{Fire Data Source}\tnote{3} &
        \textbf{Years} \\
        \midrule

        \makecell[l]{CalWildFire} & \cite{DeRango2026calwildfire} & Calabria (Italy) & Daily & \makecell{100\,m} & 8,609 & 21 & M, T, FFA, FM, FS, A & F & 2008--2018 \\

        \makecell[l]{Canadian Fire\\[-2pt]Spread Dataset} & \cite{barber_2024_canadianfirespread} & Canada & Daily & 180\,m & 3,296 & 50 & M, V, T, PFA, FM, FS, A & M, V, L, S & 2002--2021 \\
        
        \makecell[l]{Wildfire-\\[-2pt]SpreadTS+} & \cite{lahrichi_2025_wsts+} & W-Central US & Daily & \makecell{375\,m--\\[-2pt]27\,km} & 1,005 & 40 & M, V, T, FPX, FM & V, M, L & 2016--2023 \\
        
        TS-SatFire & \cite{zhao_2025_tssatfire} & CONUS & 1\,h--1\,yr & 375\,m & 179 & 27 & M, V, T, FPX & V, M & 2017--2021 \\
        
        FireSpread\_MedEU\tnote{4} & \cite{muller_high-resolution_2026} & Mediterranean \& EU & Daily & 3\,m & 103 & 21 & V, PFA, A & P & 2017--2023 \\
        
        IberFire\tnote{5} & \cite{erzibengoa_2025_iberfire} & Spain & Daily & 1--9\,km & -- & 120 & M, V, T, FPX, FS, FM, A & S & 2008--2024 \\
        
        Mesogeos & \cite{kondylatos_2023_mesogeos} & Mediterr.\ Basin & Daily & 1\,km & -- & 27 & M, V, T, FM, A, EF & M & 2006--2022 \\

        \makecell[l]{Next Day\\[-2pt]Wildfire Spread} & \cite{huot_2022_nextdaywildfirespread} & \gls{conus} & Daily & 1\,km & -- & 11 & M, V, T, FPX, FM, A & M & 2012--2020 \\

        WildfireDB & \cite{singla_2020_wildfiredb} & CONUS & Daily & 30--375\,m & -- & 20+ & M, V, T, FPX & V & 2012--2017 \\

        Wildfire-Dataset & \cite{ali_2024_dataset} & \makecell{CONUS,\\[-2pt]N.\ America} & Daily & 500\,m & -- & 8 & M, T & V & 2012--2024 \\

        BCWildfire & \cite{xu_2025_bcwildfire} & British Columbia & Daily & ${\sim}$1\,km & -- & 38 & M, V, T, A, FPX, FS, FM & M & 2000--2024 \\

        CAWFI & \cite{bhowmik_2025_cawfi} & California & Daily & 375\,m & -- & 13 & M, V, T, FS, FPX & M, V & 2012--2018 \\
        \midrule

        PyroStack & This work & CONUS and Alaska & 1--12\,h & 30\,m--9\,km & 6,994 & 27 & M, V, T, PFA, AF, FPX, FS, FM, A & V, L & 2012--2024 \\

        \bottomrule
    \end{tabularx}
    \begin{tablenotes}
    \footnotesize
        \raggedright
        \item[1] \textbf{Number of fire trajectories:} This column reports the number of distinct wildfires tracked throughout their entire lifecycle; values are omitted for datasets that report individual snapshots (e.g., fire days) or spatially gridded fire pixels rather than discrete fire events.
        \item[2] \textbf{Variable types:} Meteorology (M), Vegetation (V), Topography (T), Final Fire Area (FFA), Progressive Fire Area (PFA), Active Fireline (AF), Fire Pixel Location Data (FPX), Fuel Structure (FS), Fuel Moisture (FM), Anthropogenic Land Surface Variables (A), and Ecological Features (EF). Final fire area refers to fire areas only being reported at the final timestep or for the entire fire; progressive firea area refers to fire areas being reported at each intermediate timestep.
        \item[3] \textbf{Fire Data Source:} MODIS (M), LANDSAT (L), VIIRS (V), Sentinel-2 (S), Planet (P) and field-based (F). This column lists only sensors or products used for active fire detection or burned area mapping; sensors used solely to derive auxiliary variables (e.g., vegetation, weather, etc.) are excluded.
        \item[4] Fire perimeters are derived from 3m resolution images, but all other data layers contain information for the whole fire instead of rasterized images.
        \item[5] Fire data for the IberFire dataset \citep{erzibengoa_2025_iberfire} is captured using the Sentinel-2 satellite, but also a mixture of other data sources.
    \end{tablenotes}
    \end{threeparttable}
\end{table}

\section{Data and Methods}

\subsection{Overview}

PyroStack is a standardized multi-band spatiotemporal dataset designed to support reproducible development and evaluation of of wildfire spread models\citep{pyrostack_dataset}. In this section, we describe the structure of the dataset, including array organization, spatial and temporal resolutions, and the conventions used to ensure internal consistency. The dataset contains 6,994 fires that occurred between 2012 and 2024 across the United States, each with a final burned area exceeding 4 $\text{km}^2$ (1,000 acres). For each of these fires, the dataset contains a cube-like structure whose axes represent the two spatial dimensions (x and y) and time (t). 

PyroStack is constructed through a multi-stage data integration pipeline that combines satellite-derived fire progression maps with environmental covariates from multiple external sources. Fire progression is derived from active fire detections by the VIIRS instrument onboard the Suomi National Polar-orbiting Partnership (S-NPP) satellite using the \gls{feds} framework \citep{chen_2022_viirs}, and are constrained by \gls{mtbs} perimeters \citep{Eidenshink2007_mtbs,Picotte2020_mtbs_changes,chen2026_fedsmtbs_dataset}. Complementary layers that describe climate, fuels, and topography associated with each fire are retrieved, reprojected, and resampled from multiple sources. These heterogeneous inputs are then harmonized onto a common spatial grid and coordinate system, temporally aligned, and organized into standardized data cubes for each fire. The following sections describe this pipeline in detail, including fire tracking, layer acquisition, and the preprocessing steps used to ensure spatial and temporal consistency across all variables. The PyroStack dataset is archived on Zenodo \citep{pyrostack_dataset}, while the software used to generate the dataset, as well as Python scripts for running machine learning and physics-based models, are available on GitHub \citep{pyrostack_github}. We also provide a manifest file that contains the name, size, duration, and spatial bounding box latitude/longitude coordinates for each fire in PyroStack, along with its \gls{mtbs} identifier.

\subsection{Dataset Structure}

To facilitate data access, each fire is indexed in the dataset based on its unique identifier (Fire ID) reported in the \gls{mtbs} dataset \citep{chen2026_fedsmtbs_dataset}. For each fire, data layers are organized into five thematic categories (see Table \ref{table:layerCats}). The \texttt{fire\_spread} category contains all the gridded variables describing fire progression and behavior, including the burned area, active fireline, new active fire pixel locations, and fire radiative power (FRP). These layers originate from VIIRS satellite observations that have an original spatial resolution of 375m at nadir (growing to roughly 800m at the swath edge), and are resampled here to 300m for our dataset \citep{schroeder_2014_viirs}. The \gls{viirs} fire observations are reported twice daily at 1:30am and 1:30pm local time, corresponding to the two overpass times of the S-NPP satellite.

\begin{table}[t]
    \centering
    \scriptsize
    \setlength{\tabcolsep}{3pt}
    \renewcommand{\arraystretch}{1.15}

    \caption{Summary of dataset layer categories and data sources. Spatial and temporal resolution correspond to processed data in PyroStack, rather than the original data source.}
    \label{table:layerCats}

    \begin{tabularx}{\linewidth}{
        Y
        Y
        C{1.7cm}
        C{1.7cm}
        Y
    }
        \hline
        \textbf{Layer Category} &
        \textbf{Data Source} &
        \textbf{Spatial Resolution} &
        \textbf{Temporal Resolution} &
        \textbf{Directory Name} \\
        \hline

        Fire Characteristics &
        FEDS-MTBS \citep{chen2026_fedsmtbs_dataset} &
        300\,m &
        12-hr &
        \texttt{fire\_spread} \\

        Low Resolution Climate &
        ERA5-Land \citep{munoz-2021-era5land} &
        9000\,m &
        1-hr &
        \texttt{low\_res\_climate} \\

        High Resolution Climate &
        RTMA via CloudFire Inc. Worldgen Server &
        600\,m &
        1-hr &
        \texttt{high\_res\_climate} \\

        Fuel Structure &
        LANDFIRE via CloudFire Inc. Worldgen Server &
        30\,m &
        Static &
        \texttt{fuel\_structure} \\

        Vegetation, Fuel Model, and Topography &
        LANDFIRE \citep{ryan2013landfire} &
        30\,m &
        Static &
        \texttt{veg\_fm\_topo} \\

        \hline
    \end{tabularx}
\end{table}

The climate data are differentiated based on spatial resolution. Low resolution meteorological variables, including precipitation, surface pressure, and temperature from the \gls{ecmwf} Reanalysis version 5 (ERA5-Land), are provided at 9km spatial resolution \citep{munoz-2021-era5land}. Fuel moisture content and wind fields, which are required drivers for fire spread modeling, are derived from the National Centers for Environmental Prediction (NCEP) Real-Time Mesoscale Analysis (RTMA) product and provided at 600m spatial resolution via the CloudFire Inc. Worldgen Server \citep{depondeca_2011_noaa_analysis, cloudfire_worldgen_server}. Fuel structure data are also obtained from the CloudFire Inc. Worldgen Server and are provided with a spatial resolution of 30m \citep{cloudfire_worldgen_server}. Similarly, vegetation, fuel model, and topography data derived from the LANDFIRE data product are also provided at 30m resolution \citep{ryan2013landfire}.

The climate variables are reported at hourly time steps, whereas fire progression and fire behavior data are provided at 12-hour intervals. In contrast, the fuel structure, vegetation, fuel model, and topography layers are annual or static and are included once for each fire event. All data layers are defined on grids with spatial resolutions that are integer multiples of 30m. Using 30m as the base resolution allows coarser-resolution products to be mapped onto a common grid without introducing interpolation artifacts, ensuring that all layers remain spatially aligned on a consistent pixel lattice. Additional details and descriptions of individual variables and data layers are provided in the following sections. 

In addition to standardizing spatial resolution, we projected all spatial layers into a common \gls{crs}: EPSG:5070 (NAD83 / Conus Albers). Using a single projected CRS reduces alignment errors that arise when combining datasets in different native projections. EPSG:5070 expresses horizontal distances in meters, which is essential for our workflow. Specifically, it allows spatial padding, cropping, buffering, and grid resampling to be performed using linear units rather than angular degrees, thereby ensuring more accurate and internally consistent spatial operations. This also allows for consistency with our  spatial resolution units of meters. In addition, adopting a unified CRS allows for consistent tiling and overlay across fires and across all static and dynamic variables.

To facilitate analysis of the environmental conditions associated with fire ignition and termination, the data stack includes climate observations spanning one day before the reported fire start time and one day after the reported fire end time. This is implemented by extending the fire duration window in the event summary table, with the padded interval subsequently used in requests to retrieve both low- and high-resolution climate data. To ensure temporal consistency across datasets and avoid ambiguities associated with differing local time zones, all timestamps are converted to Coordinated Universal Time (UTC). 

Similarly, to better characterize the influence of fuels, topography, and other physical barriers on fire spread, we expanded the spatial domain of the final perimeter of each fire by 0.1$^\circ$ in all four cardinal directions before sending requests to data sources for climate, fuel structure, and land surface data layers. This buffered bounding box ensures that the downloaded datasets extend beyond the final fire perimeter and therefore fully encompass the fire perimeter at all stages of fire growth. This also reduces the likelihood of missing relevant spatial information due to coordinate rounding or uncertainties in the estimated final fire perimeters. 

\subsection{Fire Spread Data Layers}

Using the final perimeters from the \gls{mtbs} dataset \citep{Eidenshink2007_mtbs,Picotte2020_mtbs_changes,chen2026_fedsmtbs_dataset} as constraints, we ran the FEDS algorithm \citep{chen_2022_viirs} on VIIRS S-NPP observations to generate 12 hourly snapshots of fire properties for wildfires and prescribed fires in the contiguous United States and Alaska. In PyroStack, we included all MTBS (version 2024) fires that occurred between 2012 and 2024 that had a final burned area exceeding 1,000 acres. By using \gls{mtbs} final perimeters to constrain FEDS, we reduce commission errors associated with VIIRS misregistration, avoid reporting information for stationary non-vegetation fires and other thermal features, and limit errors associated with the merging of multiple fire events. We used the MTBS-reported ignition dates to define the starting time step in FEDS, and then tracked 12-hourly fire areas for each fire until no VIIRS active fire detections were observed for five consecutive days. We then post-processed the dataset to remove time steps at the end of each fire event in which the cumulative burned area exceeded 99.9\% of the final perimeter, to minimize computational artifacts from residual detections that persist after the fire has stopped growing. In addition, fires in the MTBS database were excluded for which no corresponding VIIRS active fire detections were available to constrain progression. Hawaii fires were also omitted due to unavailable environmental covariate data layers. After applying this filtering, the final PyroStack dataset included 6994 fires between 2012 and 2024. 

The FEDS-MTBS workflow described above produces multiple vector fire progression variables at 12-hour intervals that are useful for both fire spread model initialization and benchmarking. The fire area variable (\texttt{farea}) represents the cumulative area contained within the fire perimeter at each time step, serving as a primary benchmark target variable for both physics-based and machine-learning fire spread models. The active fireline variable (\texttt{fline}) is a binary pixel-level indicator at a 300m spatial resolution that identifies portions of the fire perimeter that are actively burning, based on the spatial proximity of the perimeter segments to satellite-derived active fire detections \citep{chen_2022_viirs}. This layer is particularly valuable for model initialization because it provides information on the location of the advancing fire front. Two additional layers indicate (i) the presence or absence of a satellite-derived thermal anomaly and (ii) the mean fire radiative power (FRP) of the most recent satellite-derived active fire pixel (\texttt{nfp} and \texttt{frp}). Importantly, not all areas within the perimeter of the gridded fire area contain new fire pixel count or radiative power information at a given time step. In some cases, the fire front may already have passed through an area, leaving behind burned regions that are no longer actively combusting and therefore no longer generating detectable thermal anomalies. In other cases, active burning may be obscured by smoke or cloud cover, limiting satellite detection capability.

\begin{figure}[t]
\includegraphics[width=\textwidth]{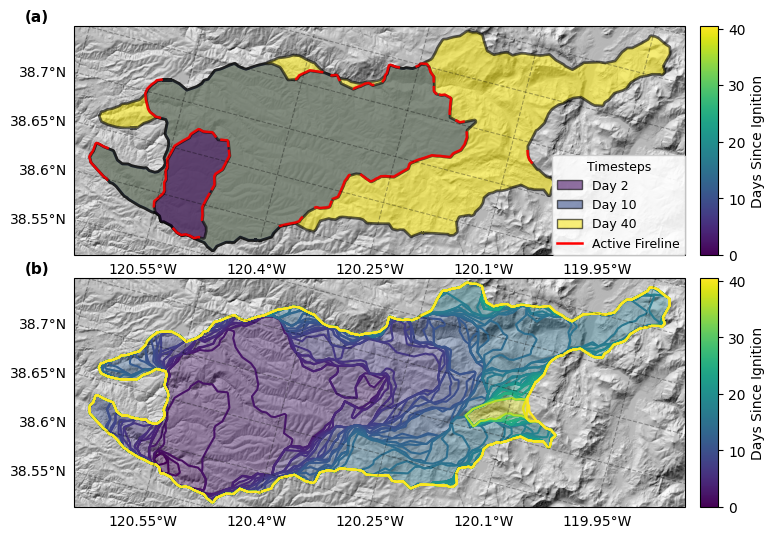}
\caption{Burn map for progression of the Caldor fire. The extent of the plot indicates the bounding box as determined by the final fire area (\texttt{farea}). Fire progression (with perimeters from FEDS-MTBS) fire areas are shown for (a) three individual timesteps corresponding to days 2, 10, and 40, with active firelines (\texttt{fline}) shown in red at each timestep, and (b) 12-hour intervals through the entire fire duration.}\label{fig:caldor_dual}
\end{figure}

To obtain spatial maps, FEDS-MTBS outputs were rasterized directly onto each fire's final PyroStack grid. For each event, we defined a common north-up grid in EPSG:5070 from the final cropped covariate extent and represented the fire-spread layers at 300 m resolution. The cumulative fire area (\texttt{farea}), active fireline (\texttt{fline}), and new fire-pixel (\texttt{nfp}) geometries were reprojected to EPSG:5070 and rasterized at each 12-hour FEDS time step, with pixels containing any portion of the original polygonal geometry assigned 1 and all other cells assigned 0. Figure \ref{fig:caldor_dual} depicts \texttt{farea} and \texttt{fline} across the full bounding box for the Caldor fire at select timesteps, as well as the progression of these layers over the entire fire duration. The \texttt{frp} layer was generated separately from the associated VIIRS active-fire pixel table: the total FRP reported for each VIIRS active-fire pixel was normalized by the reported fire-pixel footprint area to obtain a pixel-averaged FRP density, and those point values were rasterized onto the same 300 m grid.

\subsection{Low Resolution Climate Data from ERA5-Land}

ERA5-Land data \citep{era5land_dataset} are obtained through the \gls{ecmwf} Copernicus Climate Data Store API\footnote{See \url{https://cds.climate.copernicus.eu} for Climate Data Store home page.} and processed into separate raster layers for each variable\footnote{Intermediate data processing steps include obtaining hourly GRIB files per data layer, merging these into a single NetCDF dataset, then extracting and rasterizing individual variables.}. Since all spatial inputs in our dataset are standardized to a common projected grid, the ERA5-Land layers are reprojected from their native geographic coordinates to the EPSG:5070 CRS. Following reprojection, the layer data is resampled to a 9 km spatial resolution using bilinear interpolation, which is an exact multiple of our 30 m base resolution.

We produce four dynamic, hourly-resolved climate variables. This includes the 2-meter temperature (\texttt{t2m}) and dewpoint temperature (\texttt{d2m}) to represent the thermodynamic and moisture conditions of the atmospheric surface layer. These two variables can be combined to derive intermediate moisture variables relevant to fire behavior, such as relative humidity, vapor pressure deficit, and equilibrium moisture content. We also include surface pressure (\texttt{sp}) for a complete representation of the thermodynamic state of the surface atmosphere, and total precipitation (\texttt{tp}) to account for local rainfall and its influence on fuel moisture.

\subsection{High Resolution Climate Data from CloudFire Inc.}

High-resolution climate fields from Real Time Mesoscale Analysis (RTMA), including an hourly, two-dimensional analysis product of surface meteorology in the US were downloaded from the Cloudfire Inc. Worldgen server \citep{depondeca_2011_noaa_analysis, cloudfire_worldgen_server}. Since RTMA is an operational analysis rather than a frozen-version reanalysis, the underlying system underwent successive upgrades over the PyroStack study period, progressing from the legacy RTMA v2.0 framework in 2012 to RTMA v2.10 in 2023 \citep{ncep2023rtma}. While the spatial grid geometry remained consistent throughout the study period, the underlying operational background models evolved, transitioning from a downscaled 13-km Rapid Refresh (RAP) framework to a high-resolution blend of the 3-km High-Resolution Rapid Refresh (HRRR) and 4-km North American Mesoscale (NAM) models implemented in 2015. The system applies downscaling, terrain correction, and assimilation of surface observations to the operational forecasts to produce the 2.5-km analysis dataset \citep{depondeca_2011_noaa_analysis}. With the surface conditions gathered from RTMA, fuel moisture products were derived via the fuel moisture models from the National Fire Danger Rating System (NFDRS) \citep{jolly_2024_firedanger}.

 Each data request was returned as a compressed TAR file that contained multiple GeoTIFF layers. After download, individual raster layers were extracted and isolated. Although the RTMA products are provided in their own native projection, all layers were reprojected to EPSG:5070 to maintain consistency with the other spatial data downloads. Following reprojection, the climate layers were resampled to 600m using bilinear interpolation to retain a spatial resolution that is an integer multiple of 30m. One exception is the wind direction (\texttt{wd}) layer, which was resampled using nearest-neighbor interpolation. Upsampling from the native 2.5-km resolution to 600m ensures spatial continuity and mitigates coarse grid-scale artifacts, and is not intended to represent sub-kilometer scale climate dynamics.

As a result of the processing above, PyroStack stores seven variables. This includes layers for the wind field and fuel moisture content, each resolved dynamically with hourly resolution. The wind field is represented by wind speed (\texttt{ws}) and wind direction (\texttt{wd}), which are estimated at a height of 20 ft (6.1 m) above ground level. These wind products represent the ambient sustained wind speed above the forest canopy rather than sub-canopy flow. To estimate accurate sub-canopy wind speeds, the effect of the fuel structure (described in the following section) must be parameterized to extrapolate from 20 ft to the flame height at the surface \citep{albini-baughman-1979}. The fuel moisture content is represented by a set of five fuel moisture layers corresponding to five categories of fuels: dead fuels of three size classes (\texttt{m1}, \texttt{m10}, and \texttt{m100}), where \texttt{m1} designates fuels that take 1 hour to equilibrate with the atmosphere’s equilibrium moisture content; and live fuels of two types, herbaceous (\texttt{lh}) and woody (\texttt{lw}). Note that the fuel moisture layers are independent of the actual fuel type or loading at a given cell. That is, in order for the fuel moisture content at a cell to be computed, the relative fractions of constituent fuels (e.g., the fraction of 1-hour fuels) must be estimated to compute a weighted sum of the fuel moisture layers. This is typically achieved using fuel models (e.g., the \texttt{fbfm13} or \texttt{fbfm40} layers described in section \ref{subsec:veg_fm_topo}) and weighting factors, such as those defined in \cite{rothermel_1972_model}.

\subsection{Fuel Structure Data from CloudFire Inc.}

Gridded, high-resolution fuel structure variables were also downloaded in the TAR archive alongside the high-resolution climate layers from the CloudFire Inc. Worldgen server. These fuel structure layers originate from LANDFIRE and are stored at their 30m native resolution. The fuel structure layers are static (i.e. constant over time). Four fuel structure variables are included in PyroStack: canopy height (\texttt{ch}), canopy cover (\texttt{cc}), canopy base height (\texttt{cbh}), and canopy bulk density (\texttt{cbd}). These variables represent the availability and arrangement of fuels in the vertical and horizontal dimensions, which directly influence the potential for crown fires to propagate. Fuel structure also modifies the wind conditions at the surface, and is crucial for estimating mid-flame wind speeds that drive fire propagation.

\subsection{Vegetation, Fuel Model, and Topography Data from LANDFIRE} \label{subsec:veg_fm_topo}

\begin{table}[t]
    \centering
    \setlength{\tabcolsep}{4pt}
    \caption{Summary of LANDFIRE (LF) version selection across temporal ranges with their respective sources. These versions and sources were used for the retrieval of time-varying fuel layers (e.g., evt, fbfm13, fbfm40, cc, ch, cbd, cbh) and not for time-invariant layers provided by LANDFIRE (e.g., aspect, elevation, slope). In the source columns, LF and CF refer to fuel layers retrieved through the LANDFIRE API and CloudFire Inc. Worldgen server, respectively.}
    \label{table:landfire_ranges}
    \begin{tabular}{cccccccc}
        \hline
        \multicolumn{2}{c}{\textbf{Ignition Date Range}} & & & \multicolumn{2}{c}{\textbf{CONUS Source}} & \multicolumn{2}{c}{\textbf{AK Source}} \\
        \cline{1-2}
        \cline{5-8}
        \textbf{Start Date} & \textbf{End Date} & \textbf{LF Version} & \textbf{Code} & \textbf{LF} & \textbf{CF} & \textbf{LF} & \textbf{CF} \\
        \hline
        1 Oct 2024 & 31 Dec 2024 & LF 2024 & 2.5.0 & API & API & API & Archive \\
        1 Oct 2023 & 30 Sep 2024 & LF 2023 & 2.4.0 & API & API & API & Archive \\
        1 Jan 2023 & 30 Sep 2023 & LF 2022 & 2.3.0 & API & API & API & Archive \\
        1 Jan 2021 & 31 Dec 2022 & LF 2020 & 2.2.0 & Archive & API & Archive & Archive \\
        1 Jan 2017 & 31 Dec 2020 & LF 2016 & 2.0.0 & API & API & API & Archive \\
        1 Jan 2015 & 31 Dec 2016 & LF 2014 & 1.4.0 & Archive & API & Archive & Archive \\
        1 Jan 2013 & 31 Dec 2014 & LF 2012 & 1.3.0 & Archive & API & Archive & Archive \\
        1 Jan 2012 & 31 Dec 2012 & LF 2010 & 1.2.0 & Archive & API & Archive & Archive \\
        \hline
    \end{tabular}
\end{table}

To obtain fuel, vegetation, and topographic information from the LANDFIRE product suite, we use the LANDFIRE API to retrieve the appropriate layers for each fire. LANDFIRE updates its datasets after major fire events, so PyroStack assigns each fire a corresponding LANDFIRE version of its associated data based on the year preceding the fire. This ensures that the inputs represent pre-fire conditions rather than post-disturbance states that would inadvertently encode information from the fire itself. Due to the limited availability of LANDFIRE versions from the API, for fires with sub-optimal version availability, we retrieve fuel and vegetation layers directly from LANDFIRE full-extent archives (see Table \ref{table:landfire_ranges} for year-to-version mapping and a summary of the versions retrieved from archives). LANDFIRE versions change often and mapping methodology has changed across versions; thus, users should recognize that differences in versions may influence downstream analyses. To ensure full spatial coverage, we define the retrieval bounding box as the spatial extent of the final fire perimeter expanded by $0.2 ^ \circ$ in latitude and longitude in all directions. This spatial padding provides a margin that prevents gaps along the edges of the fire domain after reprojection or cropping.

LANDFIRE returns each requested dataset as a compressed ZIP archive containing GeoTIFF files. The downloaded rasters are delivered in ESRI:102003, a projected CRS (Albers Equal Area) that is similar to EPSG:5070 but not identical. The native LANDFIRE data have a spatial resolution close to, but not exactly, 30m, which additional processing and resampling for PyroStack. To maintain consistency across all layers, each LANDFIRE layer is reprojected into EPSG:5070 and resampled to a 30m resolution with nearest-neighbor interpolation. This ensures pixel-level alignment between LANDFIRE variables and other data layers. Note that all LANDFIRE layers are annual or static (i.e. constant over the time of each fire).

This LANDFIRE workflow produces multiple high-resolution land cover classification and topographic layers that are useful for both landscape categorization and as inputs for fire prediction models. The existing vegetation type layer (\texttt{evt}) is a categorical layer that maps each cell to one of hundreds of ecological systems classes, which can be further grouped into vegetation, physiognomy, and lifeform classes. The fuel model layers (\texttt{fbfm13} and \texttt{fbfm40}) are also categorical layers, which were specifically designed to generalize fuel properties into categorical inputs for operational fire behavior models. These layers map a set of fuel models characterized by fuel type, fuel loading, and climate type to a set of twelve physical variables (e.g., 1-hour fuel load, live woody fuel load, fuel bed depth) that function as direct inputs to physics-based fire models \citep{anderson_1982_fuelmodels, scott-burgan}. These fuel models are necessary for converting the fuel moisture layers to fuel-type-specific estimates of local fuel moisture content. The topographic layers are elevation (\texttt{*elev*}), slope (\texttt{*slpd*}), and aspect (\texttt{*asp*}). We use asterisks in the layer abbreviated names to indicate that different LANDFIRE versions may introduce prefixes or suffixes during data acquisition. These fields are highly influential on the direction and magnitude of fire propagation. Finally, we also include the roads mask (\texttt{*roads*}) with the aforementioned layers since it is a LANDFIRE-derived product. 

\subsection{Data Postprocessing}

After reprojection and resampling, the final step for all raster layers is to ensure that they share an identical spatial extent and are aligned on the same grid. For each fire, the spatial domain is first defined using the bounding box of the fire’s final perimeter. This bounding box is converted into the common CRS (EPSG:5070), and its center point is used as the basis for all subsequent operations. To ensure that the final spatial domain is compatible with the multi-resolution structure of the dataset, we use the coarsest spatial resolution among all layers, which is 9km (from ERA5-Land resampled data). The width and height of the bounding box is then expanded so that the final domain dimensions are exact multiples of this coarsest resolution. This guarantees that all layers can be cleanly represented on the same pixel lattice without producing fractional or misaligned grid cells.

A small minimum padding threshold is also applied. If the residual needed to reach a multiple of the coarsest resolution is less than 10\% of that resolution (i.e. less than 900m), we add a full-resolution padding increment. The resulting expanded window is then centered on the fire’s bounding-box midpoint, ensuring that the spatial domain is symmetric and consistent across all layers. Each raster is then cropped to these unified bounds. For each layer, we finally compute the corresponding pixel window while rounding boundaries upward to ensure that the crop encloses the entire intended area. 

A brief overview of each layer in the PyroStack dataset can be found in Table \ref{table:layersDescription}. Here, we provide the layer name as it appears in the dataset, a short description of the layer, its category, its original data source, and its measurement units. Note that LANDFIRE filenames vary across versions, so we use asterisks to denote the presence of additional prefixes or suffixes. In a similar manner, the layers \texttt{f13} and \texttt{f40} may also appear as \texttt{fbfm13} and \texttt{fbfm40} in PyroStack. For a detailed description of each layer, see Appendix \ref{app:layer_desc}.

\begin{table}[t]
\centering
\small
\begin{threeparttable}

\caption{Brief description of the data layers available for each fire in PyroStack.}
\label{table:layersDescription}

\setlength{\tabcolsep}{3pt}

\begin{tabularx}{\linewidth}{
    >{\ttfamily\centering\arraybackslash}p{1.5cm}  
    Y                                                 
    Y                                                
    C{2.2cm}                                         
    C{1.8cm}                                          
}
\hline
\textbf{Layer Name}\tnote{1} &
\textbf{Description} &
\textbf{Category} &
\textbf{Data Source} &
\textbf{Units} \\
\hline

farea & Fire area & Fire Characteristics & FEDS & Binary \\
fline & Active fireline & Fire Characteristics & FEDS & Binary \\
nfp & New fire pixels & Fire Characteristics & FEDS & Binary \\
frp & Fire radiative power & Fire Characteristics & FEDS & W/m$^2$ \\

d2m & 2-meter dewpoint temperature & Low-Res Climate & ERA5-Land & Kelvin (K) \\
sp & Surface pressure & Low-Res Climate & ERA5-Land & Pascal (Pa) \\
t2m & 2-meter temperature & Low-Res Climate & ERA5-Land & Kelvin (K) \\
tp & Total precipitation & Low-Res Climate & ERA5-Land & Meters/hr (m/hr) \\

lh & Live herbaceous fuel moisture & High-Res Climate & CloudFire Inc. & Percent (\%) \\
lw & Live woody fuel moisture & High-Res Climate & CloudFire Inc. & Percent (\%) \\
m1 & 1-hour dead fuel moisture & High-Res Climate & CloudFire Inc. & Percent (\%) \\
m10 & 10-hour dead fuel moisture & High-Res Climate & CloudFire Inc. & Percent (\%) \\
m100 & 100-hour dead fuel moisture & High-Res Climate & CloudFire Inc. & Percent (\%) \\
wd & 20-foot wind direction & High-Res Climate & CloudFire Inc. & Degrees ($^\circ$) \\
ws & 20-foot wind speed & High-Res Climate & CloudFire Inc. & Miles per hour (mph) \\

cbd & Canopy bulk density & Fuel Structure & CloudFire Inc. & 100 kg/m$^3$ \\
cbh & Canopy base height & Fuel Structure & CloudFire Inc. & $\text{m} \cdot 10$ \\
cc & Canopy cover & Fuel Structure & CloudFire Inc. & Percent (\%) \\
ch & Canopy height & Fuel Structure & CloudFire Inc. & $\text{m} \cdot 10$ \\

*evt* & Existing vegetation type & Veg., Fuel Model, Topo. & LANDFIRE & Categorical \\
*f[bfm]13* & 13 Anderson Fire Behavior Fuel Models (2020) & Veg., Fuel Model, Topo. & LANDFIRE & Categorical \\
*f[bfm]40* & 40 Scott and Burgan Fire Behavior Fuel Models (2020) & Veg., Fuel Model, Topo. & LANDFIRE & Categorical \\
*roads* & Operational roads & Veg., Fuel Model, Topo. & LANDFIRE & Categorical \\
*asp* & Topographic aspect & Veg., Fuel Model, Topo. & LANDFIRE & Degrees ($^\circ$) \\
*elev* & Topographic elevation & Veg., Fuel Model, Topo. & LANDFIRE & Meters (m) \\
*slpd* & Topographic slope & Veg., Fuel Model, Topo. & LANDFIRE & Degrees ($^\circ$) \\

\hline
\end{tabularx}

\begin{tablenotes}
    \setstretch{0.9}
    \footnotesize
    \raggedright
    \item[1] Asterisks in layer names indicate that in the PyroStack dataset, the corresponding TIFs may include prefixes and/or suffixes based on LANDFIRE naming conventions.
    \item[2] Note: \texttt{lh}, \texttt{lw}, \texttt{m1}, \texttt{m10}, \texttt{m100} are calculated using NFDRS4 \citep{jolly_2024_firedanger} using \gls{noaa}'s \gls{rtma} forcing data; \texttt{wd} and \texttt{ws} are gathered directly from \gls{rtma}. All data were processed at 2.5 km native resolution and subsequently resampled to 600 m.
\end{tablenotes}

\end{threeparttable}
\end{table}

\section{Results and Discussion}

\subsection{PyroStack Statistics}

\begin{table}[t]
\centering
\small
\setlength{\tabcolsep}{2.5pt}

\caption{Cumulative contribution of the largest fires to the total number of 12-hourly active fire observations, total fireline length, and total burned area. Data are grouped by fire size categories, where each category represents the top percentile of fires sorted by burned area.}
\label{table:percentContribution}

\begin{tabular}{
    c
    c
    c
    c c
    c c
    c c
}
\toprule

\multicolumn{2}{c}{\textbf{Fire Size Category}} &
\textbf{Fire Count} &
\multicolumn{2}{c}{\makecell{\textbf{Total Active}\\\textbf{Fire Obs.}}} &
\multicolumn{2}{c}{\makecell{\textbf{Total Fire}\\\textbf{Line Length}}} &
\multicolumn{2}{c}{\makecell{\textbf{Total Burned}\\\textbf{Area}}} \\

\cmidrule(lr){1-2}
\cmidrule(lr){4-5}
\cmidrule(lr){6-7}
\cmidrule(lr){8-9}

\textbf{Category} &
\textbf{Threshold} &
 &
\textbf{Count} &
\textbf{\%} &
\textbf{Length (km)} &
\textbf{\%} &
\makecell{\textbf{Area}\\\textbf{($10^3$ km$^2$)}} &
\textbf{\%} \\

\midrule

Top 1\%  & $>600\text{ km}^2$ & 70   & 4,098  & 7   & 102 & 21 & 71  & 27 \\
Top 5\%  & $>160\text{ km}^2$ & 350  & 13,600 & 24  & 235 & 48 & 150 & 58 \\
Top 10\% & $>70\text{ km}^2$  & 700  & 22,551 & 41  & 314 & 64 & 187 & 72 \\
Top 50\% & $>7\text{ km}^2$   & 3497 & 47,900 & 86  & 459 & 93 & 246 & 95 \\
All fires & --               & 6994 & 55,680 & 100 & 495 & 100 & 259 & 100 \\

\bottomrule
\end{tabular}
\end{table}

The PyroStack dataset currently contains 6,994 fires that span a wide range of sizes, durations, geographic locations, seasonal conditions, and topographic and ecological settings. In total, 55,680 12-hourly fire spread increments are recorded. The largest fires account for most of the burned area and 12-hour spread increments (Table \ref{table:percentContribution}). For example, the largest 10\% of fires, which have a size greater than 70 km$^2$, comprise 41\% of the fire spread increments, 64\% of the active fireline length segments, and 72\% of the total burned area. 

\begin{figure}[t]
\includegraphics[width=\textwidth]{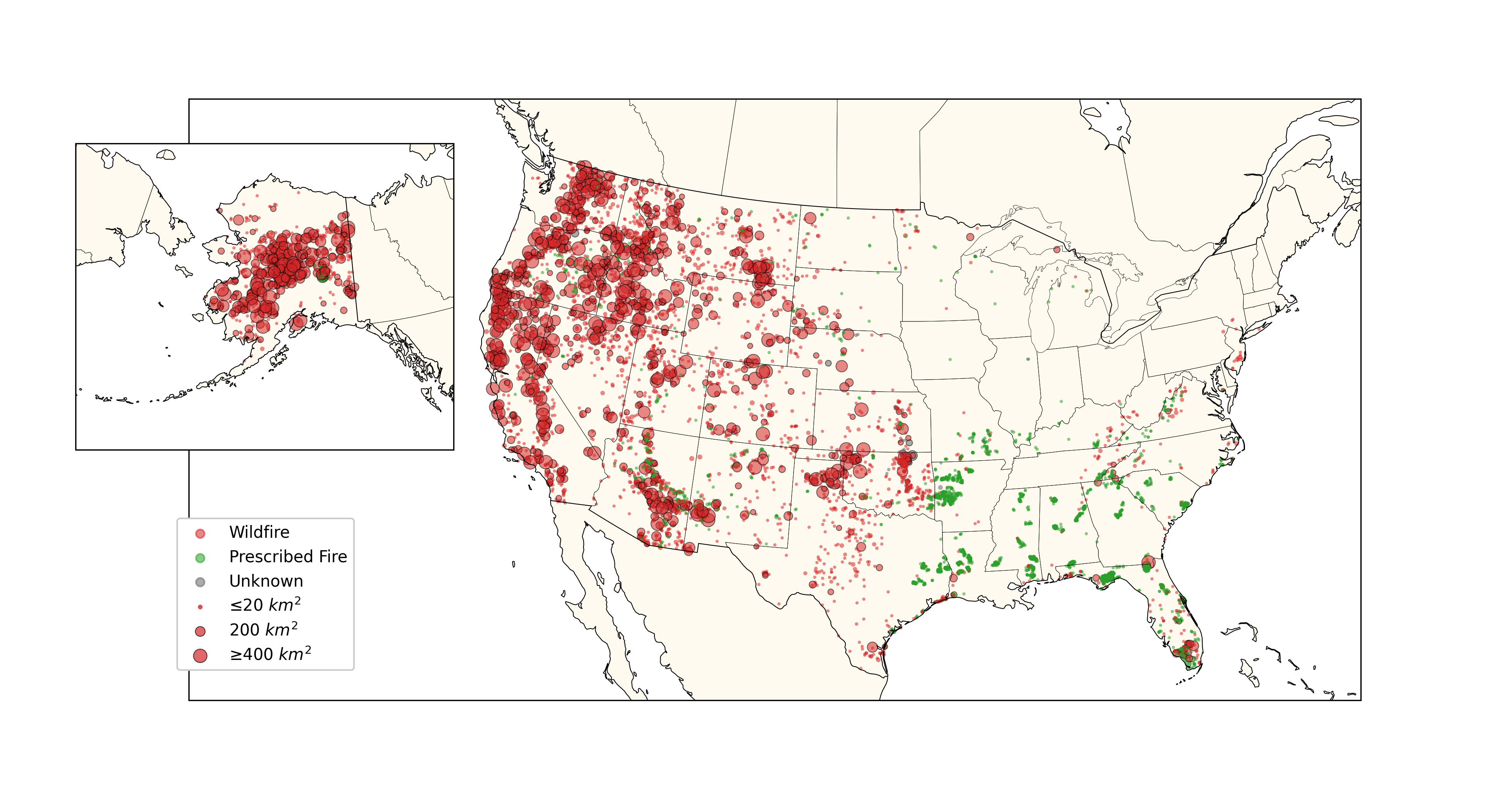}
\caption{Map of all fires included in the PyroStack dataset. Point size scales with the final burned area of each fire, while point color denotes fire type (wildfire, prescribed, or unknown). }\label{fig:spatialmap}
\end{figure}

\begin{figure}[t]
\includegraphics[width=\textwidth]{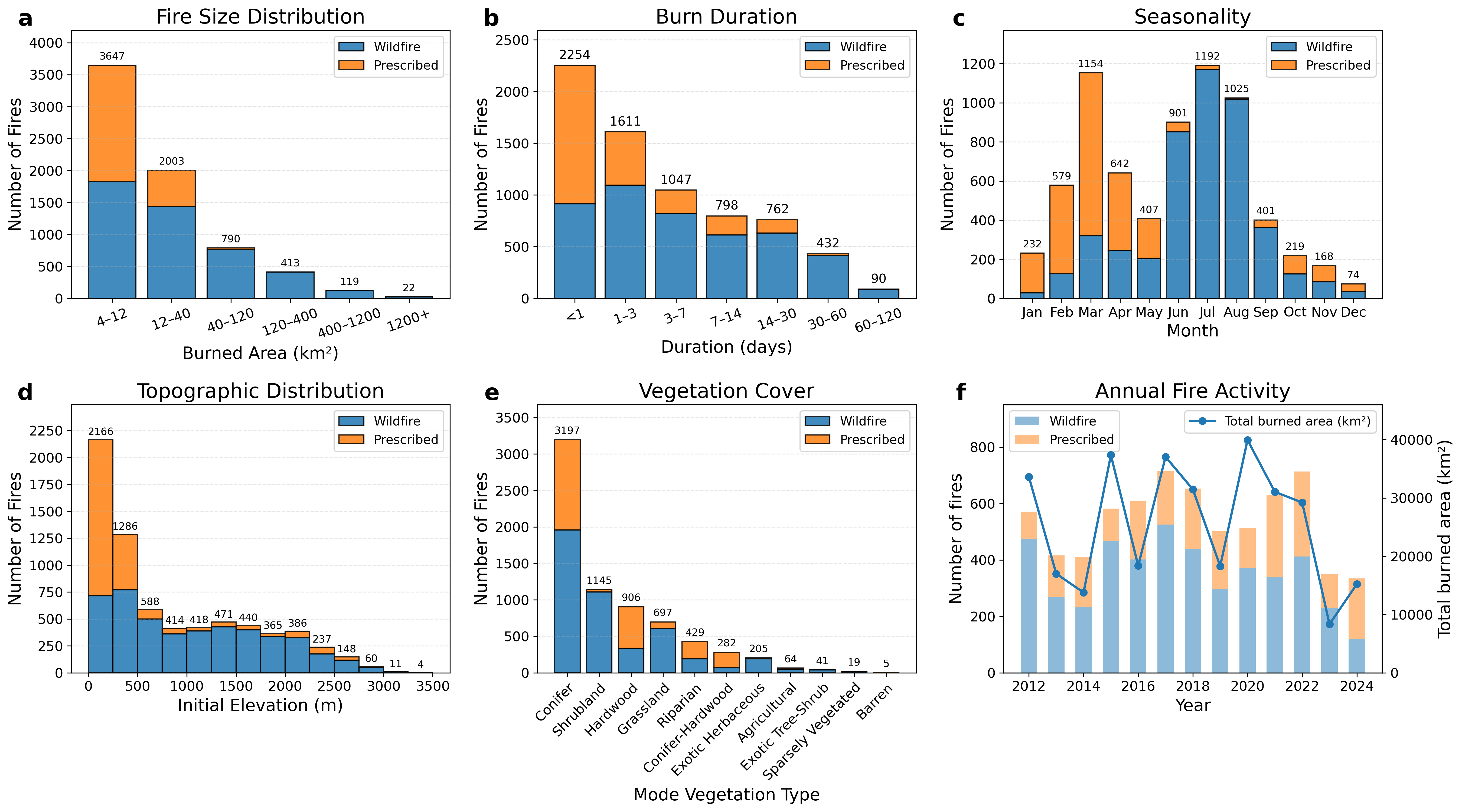}
\caption{Characteristics of wildfires and prescribed fires in PyroStack. Distributions are shown by (a) total burned area, (b) burn duration, defined as the elapsed time between the first and last FEDS time step, (c) ignition month, (d) initial elevation, (e) dominant vegetation type based on the LANDFIRE existing vegetation physiognomy product, and (f) year of occurrence. In panel (f), total annual burned area is also shown for comparison. }\label{fig:fire_summary}
\end{figure}

PyroStack's spatial and temporal domain samples wildfires and prescribed fires across multiple ecosystem types and climate regions as shown in Figure \ref{fig:spatialmap}. Wildfires are particularly prevalent in the Rocky Mountains, California, the Pacific Northwest, and interior Alaska, whereas prescribed fires are more prevalent in the Southeastern U.S.. As expected, wildfires are larger in size, have a longer duration, occur primarily during summer and early fall, are more prevalent at higher elevations, and are more common in conifer and shrubland ecosystems, as compared to prescribed fires (see Figure \ref{fig:fire_summary}). There is considerable interannual variability in the number of fires and burned area contained in the dataset, with the greatest amount of burned area occurring in 2020, a record-breaking high fire year in the Western US (see Figure \ref{fig:fire_summary}f).

\subsection{A Specific Fire Example: the Caldor Fire}

\begin{figure}[t]
\includegraphics[width=\textwidth]{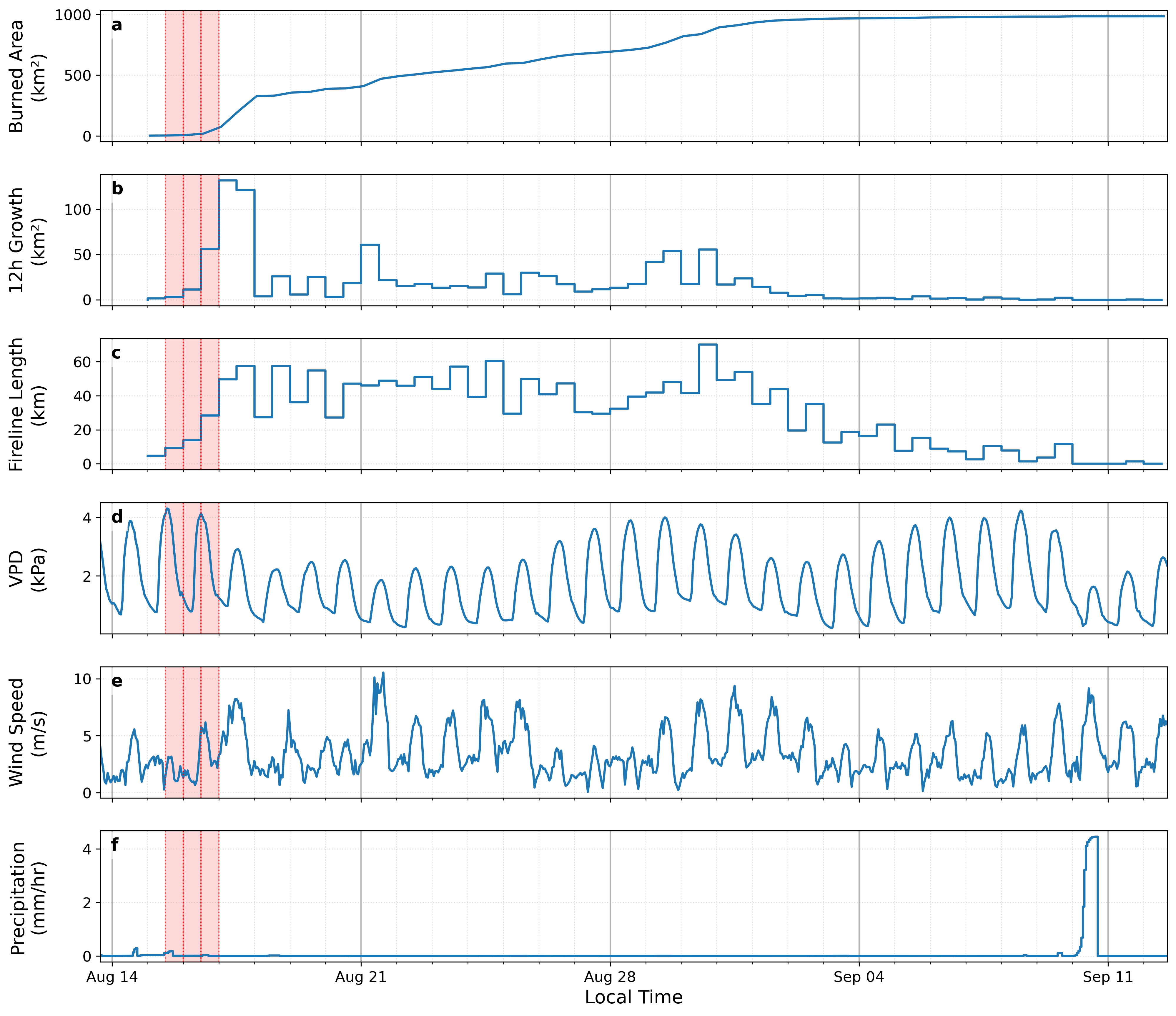}
\caption{Time series of selected variables from PyroStack for the Caldor Fire, which burned in 2021 in California’s Sierra Nevada. Panels show (a) cumulative burned area, (b) incremental burned area, and (c) active fireline length, together with selected hourly environmental drivers aggregated over the fire domain: (d) vapor pressure deficit, derived from the PyroStack temperature and dewpoint temperature layers; (e) wind speed; and (f) precipitation. The red-shaded intervals on 15–16 August 2021 indicate the three 12-hour periods used to initialize and evaluate the machine-learning- and physics-based fire-spread simulations described in Sections 3.3 and 3.4 (see Figure \ref{fig:comparison}).}\label{fig:caldor_timeseries}
\end{figure}

\begin{figure}[t]
\includegraphics[width=\textwidth]{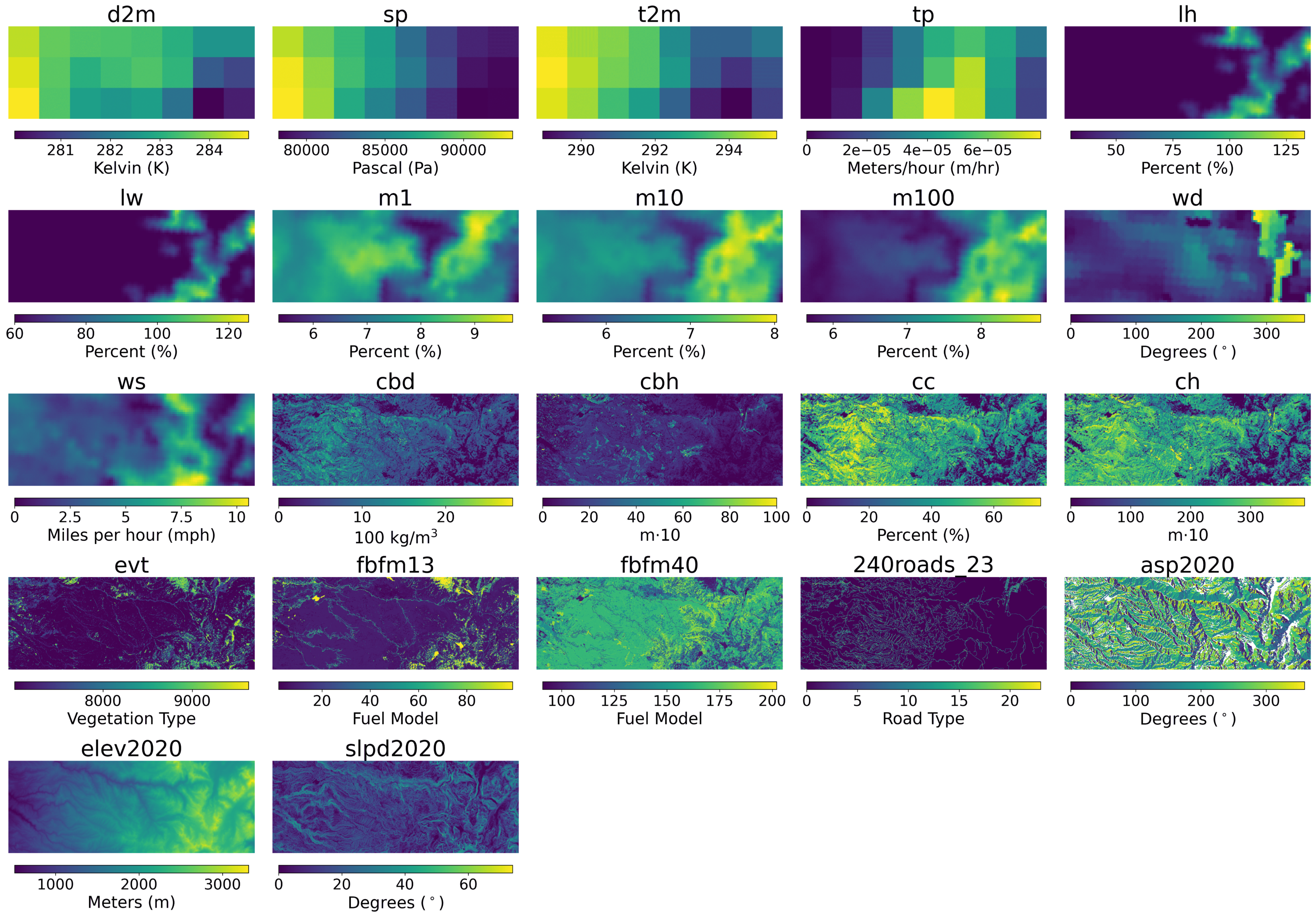}
\caption{Spatial distribution of PyroStack covariate layers (i.e. non-fire-spread layers) for the Caldor Fire. Data correspond to observations at 12PM PST on August 15, 2021.}\label{fig:caldor_tifs}
\end{figure}

The Caldor Fire was a massive and destructive wildfire that burned 221,835 acres in California, and was first reported on August 14th of the 2021 wildfire season. Anomalous fire behavior contributed significantly to the spread of the Caldor Fire, including rapid and destructive crown-fire runs and long-range firebrand lofting. These phenomena were driven by a combination of extreme fire weather and historically high fuel loads. Collectively, its extreme fire behavior and complex environmental interactions make the Caldor Fire an ideal example for demonstrating PyroStack's spatiotemporal reconstruction of a single fire. The progression of the Caldor Fire is captured by the PyroStack's burned area increments shown in Figure \ref{fig:caldor_dual}b. The period of rapid fire growth between August 15 and 17 coincided with elevated vapor pressure deficits and increased wind speeds (Figure \ref{fig:caldor_timeseries}). Vertical bars in the figure denote the timing of increments used to illustrate the spatial distribution of data layers stored within PyroStack in Figure \ref{fig:caldor_tifs}. In the following sections, we use the Caldor Fire as a test case to demonstrate the operability of the PyroStack dataset to facilitate the simulation of a fire with both an ML and a physical model.

\subsection{Machine Learning and Physics-Based Fire Spread Model Execution on PyroStack}

A key goal of PyroStack is to enable systematic benchmarking of ML and physics-based models for fire spread prediction at a large scale. For ML modeling approaches, we illustrate this capability by training a Vision Transformer (ViT) model (a widely used form of deep learning model for image analysis \citep{dosovitskiy2021an}), to predict the next-day fire perimeter. We provide a PyTorch \citep{pytorch} dataloader implementation to facilitate community adoption (see GitHub repository for Python script).

Given the observed fire state and environmental conditions at time $t$, the prediction target is the fire-area boundary (\texttt{farea}) at $t+12$ hours, at a 300m resolution. Since \texttt{farea} is a binary variable indicating whether each grid cell lies within the fire area (1) or not (0), we use a classification-based cross-entropy loss as the objective function to train the model. The model is conditioned on two sets of inputs. The first comprises the four 12-hourly fire progression variables at time $t$ - \texttt{fline}, \texttt{farea}, \texttt{nfp}, and \texttt{frp} — which characterize the current fire extent and the location of actively burning locations and the fire radiative power of these observations. The second set of inputs comprises all hourly RTMA, ERA5-Land, and LANDFIRE fields over the $[t-12\text{h}, t]$ window, providing spatially explicit information on fuel moisture, meteorological forcing, topography, and vegetation structure. Together, these inputs span the primary physical drivers of short-term fire spread.

The ViT model was trained on a set of 5,117 fires from PyroStack. The Caldor Fire was withheld from the training dataset to provide a qualitative illustration of the predicted spread probabilities on an unseen event. For a more detailed description of this model, see Appendix \ref{app:vit_details}. In parallel, to demonstrate the application of the PyroStack dataset for deploying and validating a physics-based model, we used the open-source Pyretechnics fire behavior model \citep{PyretechnicsLibrary} as a representative example. Pyretechnics employs the standard physical parameterizations used in operational models \citep{rothermel_1972_model, rothermel_1991_rockymtfirs}, implementing them within an Eulerian level-set method \citep{lautenberger_2013_ELMFIRE} for tracking fire-front propagation on a high-resolution grid. The model is designed to be initialized from ignition cells such as the FEDS active fireline rasters, and to operate on the standard fuel, topography, and climate inputs supplied in PyroStack. It can also be rapidly deployed in an ensemble of simulations with perturbed input parameters to produce a probabilistic prediction of fire spread. An example script to run Pyretechnics on the Caldor Fire is provided in our GitHub repository \citep{pyrostack_github}.

\begin{figure}[t]
\includegraphics[width=\textwidth]{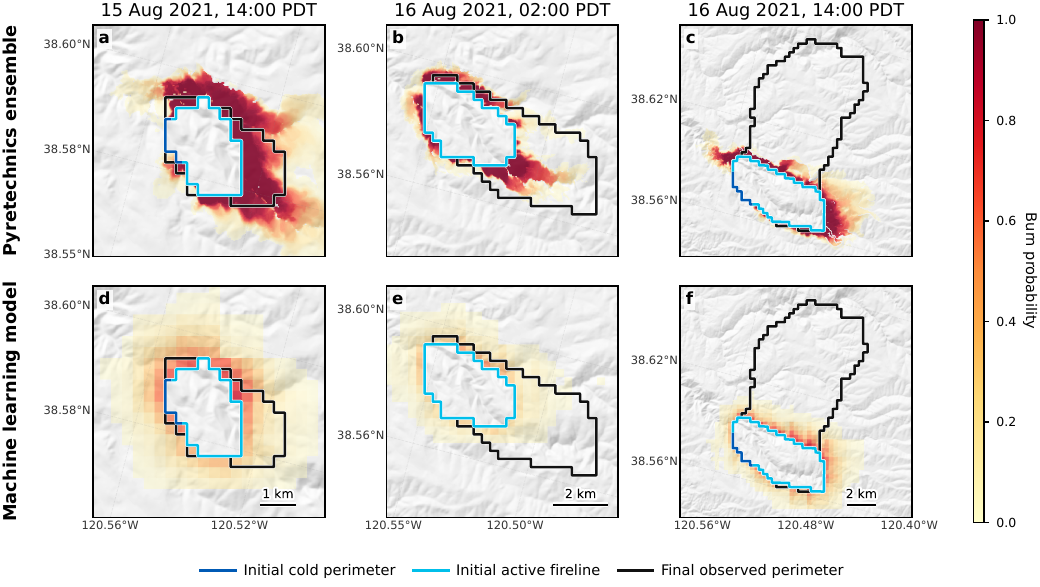}
\caption{Comparison of Pyretechnics ensemble and our machine learning model on fire spread forecasts for the Caldor Fire. Each panel shows predicted burn probability overlaid on terrain shading, with the initial cold perimeter (dark blue), initial active fireline (cyan), and observed final FEDS-MTBS perimeter (black) shown for reference. The cold perimeter distinguishes the inactive segments of the initial FEDS-MTBS perimeter from the active fireline at time $t$, while the final FEDS-MTBS perimeter represents the updated observed fire extent at time $t+12$. Panels (a, d) show forecasts initialized at 14:00 PDT on 15 August 2021, panels (b, e) show forecasts initialized at 02:00 PDT on 16 August 2021, and panels (c, f) show forecasts initialized at 14:00 PDT on 16 August 2021. The Pyretechnics ensemble (a, b, c) produces spatially concentrated, high-confidence predictions closely tied to the active fireline, reflecting its physics-based propagation rules. The ViT (d, e, f) produces smoother, more spatially diffuse probability fields, reflecting the model's tendency to spread probability mass over a broader region.
}\label{fig:comparison}
\end{figure}

To evaluate the performance of both the ML and Pyretechnics models over the minimum forecast horizon, we applied each model in a one-step-ahead prediction setting, initialized from the FEDS active fireline at time $t$. Pyretechnics was integrated forward at sub-hourly timesteps until the next available satellite-derived fire area observation 12 hours later, while the ViT directly predicts \texttt{farea} at $t + 12$ hours. For qualitative evaluation, we plot the predicted burn probability field against the observed final perimeter. We illustrate the results for three forecast periods during the rapid-growth phase of the Caldor fire; see the red shading in Figure \ref{fig:caldor_timeseries} for their timing. In Figure \ref{fig:comparison}, we display the ML and Pyretechnics model predictions for both forecast periods, demonstrating the relative capabilities of each model under challenging spread conditions. 

\section{Conclusions}

In this work, we introduced PyroStack, a large-scale, standardized spatiotemporal dataset of \gls{us} wildfires designed to both develop predictive models and improve our understanding of wildfire spread. By integrating satellite-derived fire observations with high-resolution meteorological, fuel, and topographic data, PyroStack addresses a critical gap in the availability of harmonized, analysis-ready datasets for wildfire modeling. The dataset spans nearly 7,000 fires from 2012 to 2024 and provides a unified representation of fire progression and environmental drivers on a consistent spatial grid and temporal framework.

Compared to existing datasets, PyroStack provides extensive spatial coverage, higher temporal resolution, and a richer set of variables, including 12-hour fire area and active fireline observations and 1-hour meteorological driver variables. These features enable support for a wider range of applications, from next-step fire spread prediction to the calibration and validation of semi-empirical and physics-based models. The relatively large spatial domain and time span capture wildfires and prescribed fires across many ecoregions, which is essential for identifying model strengths and limitations and advancing the state of wildfire prediction. Furthermore, it enables direct pixel-level analysis of interactions among fuels, weather, and fire dynamics. PyroStack's design facilitates reproducible workflows and lowers the barrier to entry for researchers seeking to develop and evaluate wildfire spread models. 

The availability of a comprehensive, well-documented dataset also has the potential to foster greater engagement from the machine learning community, where progress has historically been driven by access to large, standardized benchmarks. By providing an open and extensible data resource, PyroStack creates new opportunities for the development of data-driven and hybrid modeling approaches, as well as for rigorous comparisons between machine learning and traditional physical models.

Despite its broad coverage and standardized design, PyroStack has several limitations that should be considered. Uncertainties in satellite-based fire observations, including missed detections due to smoke or cloud cover, can affect the accuracy of fire progression layers. The 12-hour temporal resolution of fire observations may not fully capture rapid changes in fire behavior, particularly under extreme conditions. Additionally, although the dataset harmonizes multiple data sources with differing native resolutions and uncertainties to a common grid, reprojection and resampling may smooth fine-scale variability, especially for coarser inputs such as ERA5-Land meteorological variables. Finally, important drivers of fire behavior, such as human suppression activities, ignition processes, and fine-scale fuel heterogeneity, are not explicitly represented, which may limit certain applications. Future work will focus on addressing these limitations by expanding the data set to incorporate additional constraints, including information on fire suppression gathered by incident command for individual fires (ICS-209 reports) and other data sets on fuel and vegetation composition. We anticipate that PyroStack will serve as a foundation for community-driven efforts to advance wildfire science, ultimately contributing to improved forecasting capabilities and more effective mitigation of wildfire impacts under a changing climate.

\bibliographystyle{plainnat}
\bibliography{references}  

@article{Vitolo_2020_era5,
  title = {{ERA5}-based global meteorological wildfire danger maps},
  volume = {7},
  ISSN = {2052-4463},
  url = {https://doi.org/10.1038/s41597-020-0554-z},
  pages = {216},
  journal = {Scientific Data},
  publisher = {Springer Science and Business Media LLC},
  author = {Vitolo,  Claudia and Di Giuseppe,  Francesca and Barnard,  Christopher and Coughlan,  Ruth and San-Miguel-Ayanz,  Jesus and Libertá,  Giorgio and Krzeminski,  Blazej},
  year = {2020},
}

@article{chen_wildfire_2024,
	title = {Wildfire risk for global wildland–urban interface areas},
	volume = {7},
	issn = {2398-9629},
	url = {https://doi.org/10.1038/s41893-024-01291-0},
	pages = {474--484},
	number = {4},
    year={2024},
	journal = {Nature Sustainability},
	shortjournal = {Nature Sustainability},
	author = {Chen, Bin and Wu, Shengbiao and Jin, Yufang and Song, Yimeng and Wu, Chao and Venevsky, Sergey and Xu, Bing and Webster, Chris and Gong, Peng},
}

@article{schug_global_2023,
	title = {The global wildland–urban interface},
	volume = {621},
	issn = {1476-4687},
	url = {https://doi.org/10.1038/s41586-023-06320-0},
	pages = {94--99},
	number = {7977},
    year={2023},
	journal = {Nature},
	shortjournal = {Nature},
	author = {Schug, Franz and Bar-Massada, Avi and Carlson, Amanda R. and Cox, Heather and Hawbaker, Todd J. and Helmers, David and Hostert, Patrick and Kaim, Dominik and Kasraee, Neda K. and Martinuzzi, Sebastián and Mockrin, Miranda H. and Pfoch, Kira A. and Radeloff, Volker C.},
}

@article{jones_global_2024,
  title={Global rise in forest fire emissions linked to climate change in the extratropics},
  author={Matthew W. Jones  and Sander Veraverbeke  and Niels Andela  and Stefan H. Doerr  and Crystal Kolden  and Guilherme Mataveli  and M. Lucrecia Pettinari  and Corinne Le Quéré  and Thais M. Rosan  and Guido R. van der Werf  and Dave van Wees  and John T. Abatzoglou},
  journal={Science},
  url = {https://doi.org/10.1126/science.adl5889},
  volume={386},
  number={6719},
  pages={eadl5889},
  year={2024},
  publisher={American Association for the Advancement of Science}
}

@article{ryan2013landfire,
  title = {{LANDFIRE} – {A} national vegetation/fuels data base for use in fuels treatment, restoration, and suppression planning},
  volume = {294},
  ISSN = {0378-1127},
  url = {https://doi.org/10.1016/j.foreco.2012.11.003},
  journal = {Forest Ecology and Management},
  publisher = {Elsevier BV},
  author = {Ryan,  Kevin C. and Opperman,  Tonja S.},
  year = {2013},
  pages = {208–216}
}

@article{Scholten_2024_environment,
  title = {Spatial variability in {A}rctic–boreal fire regimes influenced by environmental and human factors},
  volume = {17},
  ISSN = {1752-0908},
  url = {https://doi.org/10.1038/s41561-024-01505-2},
  number = {9},
  journal = {Nature Geoscience},
  publisher = {Springer Science and Business Media LLC},
  author = {Scholten,  Rebecca C. and Veraverbeke,  Sander and Chen,  Yang and Randerson,  James T.},
  year = {2024},
  pages = {866–873}
}

@article{Yang2026biodiversity,
  title = {Wildfire risk for species under climate change},
  ISSN = {1758-6798},
  url = {https://doi.org/10.1038/s41558-026-02600-5},
  journal = {Nature Climate Change},
  publisher = {Springer Science and Business Media LLC},
  author = {Yang,  Xiaoye and Urban,  Mark C. and Su,  Bo and Zhong,  Ziqian and Wu,  Chao and Chen,  Deliang},
  year = {2026},
  pages = {613-621},
  volume = {16}
}

@article{Gallo2025climatechange,
  title = {Future impacts of climate change on global fire weather: {Insight} from weighted {CMIP6} multimodel ensembles},
  volume = {38},
  ISSN = {1520-0442},
  url = {https://doi.org/10.1175/JCLI-D-24-0540.1},
  number = {22},
  journal = {Journal of Climate},
  publisher = {American Meteorological Society},
  author = {Gallo,  Carolina and Dieppois,  Bastien and Quilcaille,  Yann and Chiriacò,  Maria Vincenza and Fulé,  Peter Z. and Drobyshev,  Igor and San-Miguel-Ayanz,  Jesús and Blackett,  Matthew and Eden,  Jonathan M.},
  year = {2025},
  pages = {6445–6462}
}

@article{DeRango2026calwildfire,
  title = {{CalWildFire}: {A} high-resolution in situ dataset for wildfire analysis and modeling in the {Mediterranean} region},
  ISSN = {2052-4463},
  url = {https://doi.org/10.1038/s41597-026-07212-4},
  journal = {Scientific Data},
  publisher = {Springer Science and Business Media LLC},
  author = {De Rango,  Alessio and Lo Scudo,  Fabrizio and Furnari,  Luca and Senatore,  Alfonso and D’Ambrosio,  Donato and Mendicino,  Giuseppe and Greco,  Gianluigi},
  year = {2026},
  volume = {13},
  pages = {859},
}

@article{muller_high-resolution_2026,
	title = {A high-resolution spatiotemporal wildfire propagation dataset for the {Mediterranean} and {Europe}},
	volume = {13},
	issn = {2052-4463},
	url = {https://doi.org/10.1038/s41597-026-06965-2},
    pages = {389},
	journal = {Scientific Data},
	shortjournal = {Scientific Data},
	author = {Müller, Simon and Hofmann-Böllinghaus, Anja and Chen, Zhimin and Vogel, Kristin and Benner, Philipp},
	year = {2026},
}

@misc{era5land_dataset,
  url = {https://cds.climate.copernicus.eu/doi/10.24381/cds.e2161bac},
  author = {{Copernicus Climate Change Service}},
  title = {{ERA5-Land hourly data from 1950 to present}},
  publisher = {Copernicus Climate Change Service (C3S) Climate Data Store (CDS)},
  year = {2019}
}

@inproceedings{sahoo2024simple,
 author = {Sahoo, Subham Sekhar and Arriola, Marianne and Schiff, Yair and Gokaslan, Aaron and Marroquin, Edgar and Chiu, Justin T and Rush, Alexander and Kuleshov, Volodymyr},
 booktitle = {{Advances in Neural Information Processing Systems}},
 url = {https://doi.org/10.52202/079017-4135},
 editor = {A. Globerson and L. Mackey and D. Belgrave and A. Fan and U. Paquet and J. Tomczak and C. Zhang},
 pages = {130136--130184}, 
 title = {Simple and effective masked diffusion language models},
 volume = {37},
 year = {2024}
}

@inproceedings{shi2024simplified,
 author = {Shi, Jiaxin and Han, Kehang and Wang, Zhe and Doucet, Arnaud and Titsias, Michalis},
 booktitle = {{Advances in Neural Information Processing Systems}},
 url = {https://doi.org/10.52202/079017-3277},
 editor = {A. Globerson and L. Mackey and D. Belgrave and A. Fan and U. Paquet and J. Tomczak and C. Zhang},
 pages = {103131--103167}, 
 title = {Simplified and generalized masked diffusion for discrete data},
 volume = {37},
 year = {2024}
}

@inproceedings{he2022masked,
  author={He, Kaiming and Chen, Xinlei and Xie, Saining and Li, Yanghao and Doll{\'a}r, Piotr and Girshick, Ross},
  booktitle={{2022 IEEE/CVF Conference on Computer Vision and Pattern Recognition (CVPR)}}, 
  title={Masked autoencoders are scalable vision learners}, 
  year={2022},
  pages={15979-15988},
  url={https://doi.org/10.1109/CVPR52688.2022.01553}
}

@inbook{pytorch,
    author = {Paszke, Adam and Gross, Sam and Massa, Francisco and Lerer, Adam and Bradbury, James and Chanan, Gregory and Killeen, Trevor and Lin, Zeming and Gimelshein, Natalia and Antiga, Luca and Desmaison, Alban and K\"{o}pf, Andreas and Yang, Edward and DeVito, Zach and Raison, Martin and Tejani, Alykhan and Chilamkurthy, Sasank and Steiner, Benoit and Fang, Lu and Bai, Junjie and Chintala, Soumith},
    title = {{PyTorch}: an imperative style, high-performance deep learning library},
    year = {2019},
    publisher = {Curran Associates Inc.},
    booktitle = {{Proceedings of the 33rd International Conference on Neural Information Processing Systems}},
    articleno = {721},
    numpages = {12},
    pages={8026-8037},
    url={https://dl.acm.org/doi/10.5555/3454287.3455008}
    }

@inproceedings{
dosovitskiy2021an,
title={An image is worth 16x16 words: {T}ransformers for image recognition at scale},
author={Alexey Dosovitskiy and Lucas Beyer and Alexander Kolesnikov and Dirk Weissenborn and Xiaohua Zhai and Thomas Unterthiner and Mostafa Dehghani and Matthias Minderer and Georg Heigold and Sylvain Gelly and Jakob Uszkoreit and Neil Houlsby},
booktitle={{International Conference on Learning Representations}},
year={2021},
url={https://openreview.net/forum?id=YicbFdNTTy}
}

@article{oliveira_2023_near,
	title = {A near real-time web-system for predicting fire spread across the {Cerrado} biome},
	volume = {13},
	issn = {2045-2322},
	url = {https://doi.org/10.1038/s41598-023-30560-9},
	pages = {4829},
	number = {1},
	journal = {Scientific Reports},
	shortjournal = {Scientific Reports},
	author = {Oliveira, Ubirajara and Soares-Filho, Britaldo and Rodrigues, Hermann and Figueira, Danilo and Gomes, Leticia and Leles, William and Berlinck, Christian and Morelli, Fabiano and Bustamante, Mercedes and Ometto, Jean and Miranda, Heloísa},
    year = {2023}
}

@article{hersbach_2020_era5,
author = {Hersbach, Hans and Bell, Bill and Berrisford, Paul and Hirahara, Shoji and Horányi, András and Muñoz-Sabater, Joaquín and Nicolas, Julien and Peubey, Carole and Radu, Raluca and Schepers, Dinand and Simmons, Adrian and Soci, Cornel and Abdalla, Saleh and Abellan, Xavier and Balsamo, Gianpaolo and Bechtold, Peter and Biavati, Gionata and Bidlot, Jean and Bonavita, Massimo and De Chiara, Giovanna and Dahlgren, Per and Dee, Dick and Diamantakis, Michail and Dragani, Rossana and Flemming, Johannes and Forbes, Richard and Fuentes, Manuel and Geer, Alan and Haimberger, Leo and Healy, Sean and Hogan, Robin J. and Hólm, Elías and Janisková, Marta and Keeley, Sarah and Laloyaux, Patrick and Lopez, Philippe and Lupu, Cristina and Radnoti, Gabor and de Rosnay, Patricia and Rozum, Iryna and Vamborg, Freja and Villaume, Sebastien and Thépaut, Jean-Noël},
title = {The {ERA5} global reanalysis},
journal = {Quarterly Journal of the Royal Meteorological Society},
volume = {146},
number = {730},
pages = {1999-2049},
url = {https://doi.org/10.1002/qj.3803},
year = {2020}
}

@article{reeves_2009_landfire,
    author = {Reeves, Matthew C. and Ryan, Kevin C. and Rollins, Matthew G. and Thompson, Thomas G.},
    title = {Spatial fuel data products of the {LANDFIRE} Project},
    journal = {International Journal of Wildland Fire},
    volume = {18},
    number = {3},
    pages = {250-267},
    year = {2009},
    issn = {1049-8001},
    url = {https://doi.org/10.1071/WF08086},
}

@article{wang_2021_economic,
  title = {Economic footprint of {California} wildfires in 2018},
  volume = {4},
  url = {https://doi.org/10.1038/s41893-020-00646-7},
  number = {3},
  journal = {Nature Sustainability},
  author = {Wang,  Daoping and Guan,  Dabo and Zhu,  Shupeng and Kinnon,  Michael Mac and Geng,  Guannan and Zhang,  Qiang and Zheng,  Heran and Lei,  Tianyang and Shao,  Shuai and Gong,  Peng and Davis,  Steven J.},
  year = {2021},
  pages = {252–260}
}

@article{johnston_2012_mortality,
  title = {Estimated global mortality attributable to smoke from landscape fires},
  volume = {120},
  url = {https://doi.org/10.1289/ehp.1104422},
  number = {5},
  journal = {Environmental Health Perspectives},
  author = {Johnston,  Fay H. and Henderson,  Sarah B. and Chen,  Yang and Randerson,  James T. and Marlier,  Miriam and DeFries,  Ruth S. and Kinney,  Patrick and Bowman,  David M.J.S. and Brauer,  Michael},
  year = {2012},
  pages = {695–701}
}

@article{xu_2023_exposure,
  title = {Global population exposure to landscape fire air pollution from 2000 to 2019},
  volume = {621},
  url = {https://doi.org/10.1038/s41586-023-06398-6},
  number = {7979},
  journal = {Nature},
  author = {Xu,  Rongbin and Ye,  Tingting and Yue,  Xu and Yang,  Zhengyu and Yu,  Wenhua and Zhang,  Yiwen and Bell,  Michelle L. and Morawska,  Lidia and Yu,  Pei and Zhang,  Yuxi and Wu,  Yao and Liu,  Yanming and Johnston,  Fay and Lei,  Yadong and Abramson,  Michael J. and Guo,  Yuming and Li,  Shanshan},
  year = {2023},
  pages = {521–529}
}

@article{maciasfauria_2010_climate,
  title = {Predicting climate change effects on wildfires requires linking processes across scales},
  volume = {2},
  ISSN = {1757-7799},
  url = {https://doi.org/10.1002/wcc.92},
  number = {1},
  journal = {WIREs Climate Change},
  publisher = {Wiley},
  author = {Macias Fauria,  Marc and Michaletz,  Sean T. and Johnson,  Edward A.},
  year = {2010},
  pages = {99–112}
}

@techreport{finney_1998_farsite,
  author      = {Finney, Mark A.},
  title       = {{FARSITE}: {F}ire Area Simulator---Model Development and Evaluation},
  institution = {U.S. Department of Agriculture, Forest Service, Rocky Mountain Research Station},
  year        = {1998},
  type        = {Research Paper},
  number      = {RMRS-RP-4},
  address     = {Ogden, UT},
  pages       = {47},
  url         = {https://doi.org/10.2737/RMRS-RP-4}
}

@article{finney_2011_ensemble,
  title = {A method for ensemble wildland fire simulation},
  volume = {16},
  ISSN = {1573-2967},
  url = {https://doi.org/10.1007/s10666-010-9241-3},
  number = {2},
  journal = {Environmental Modeling \& Assessment},
  publisher = {Springer Science and Business Media LLC},
  author = {Finney,  Mark A. and Grenfell,  Isaac C. and McHugh,  Charles W. and Seli,  Robert C. and Trethewey,  Diane and Stratton,  Richard D. and Brittain,  Stuart},
  year = {2011},
  pages = {153–167}
}

@article{kelso_2015_aus_simulator,
  title = {Techniques for evaluating wildfire simulators via the simulation of historical fires using the {Australis} simulator},
  volume = {24},
  ISSN = {1448-5516},
  url = {https://doi.org/10.1071/WF14047},
  number = {6},
  journal = {International Journal of Wildland Fire},
  publisher = {CSIRO Publishing},
  author = {Kelso,  Joel K. and Mellor,  Drew and Murphy,  Mary E. and Milne,  George J.},
  year = {2015},
  pages = {784–797}
}

@article{jain_2020_ml_review,
  title = {A review of machine learning applications in wildfire science and management},
  volume = {28},
  ISSN = {1208-6053},
  url = {https://doi.org/10.1139/er-2020-0019},
  number = {4},
  journal = {Environmental Reviews},
  publisher = {Canadian Science Publishing},
  author = {Jain,  Piyush and Coogan,  Sean C.P. and Subramanian,  Sriram Ganapathi and Crowley,  Mark and Taylor,  Steve and Flannigan,  Mike D.},
  year = {2020},
  pages = {478–505}
}

@article{barnes_1998_modis,
  author={Barnes, W.L. and Pagano, T.S. and Salomonson, V.V.},
  journal={IEEE Transactions on Geoscience and Remote Sensing}, 
  title={Prelaunch characteristics of the {Moderate Resolution Imaging Spectroradiometer (MODIS) on EOS-AM1}}, 
  year={1998},
  volume={36},
  number={4},
  pages={1088-1100},
  url = {https://doi.org/10.1109/36.700993}
}

@article{schroeder_2014_viirs,
  title = {The {New VIIRS} 375 m active fire detection data product: {Algorithm} description and initial assessment},
  volume = {143},
  ISSN = {0034-4257},
  url = {https://doi.org/10.1016/j.rse.2013.12.008},
  journal = {Remote Sensing of Environment},
  publisher = {Elsevier BV},
  author = {Schroeder,  Wilfrid and Oliva,  Patricia and Giglio,  Louis and Csiszar,  Ivan A.},
  year = {2014},
  pages = {85–96}
}

@article{artes_2019_dataset,
  title = {A global wildfire dataset for the analysis of fire regimes and fire behaviour},
  volume = {6},
  ISSN = {2052-4463},
  url = {https://doi.org/10.1038/s41597-019-0312-2},
  pages = {296},
  journal = {Scientific Data},
  publisher = {Springer Science and Business Media LLC},
  author = {Artés,  Tomàs and Oom,  Duarte and de Rigo,  Daniele and Durrant,  Tracy Houston and Maianti,  Pieralberto and Libertà,  Giorgio and San-Miguel-Ayanz,  Jesús},
  year = {2019},
}

@article{chen_2022_viirs,
  title = {California wildfire spread derived using {VIIRS} satellite observations and an object-based tracking system},
  volume = {9},
  ISSN = {2052-4463},
  url = {https://doi.org/10.1038/s41597-022-01343-0},
  pages = {249},
  journal = {Scientific Data},
  publisher = {Springer Science and Business Media LLC},
  author = {Chen,  Yang and Hantson,  Stijn and Andela,  Niels and Coffield,  Shane R. and Graff,  Casey A. and Morton,  Douglas C. and Ott,  Lesley E. and Foufoula-Georgiou,  Efi and Smyth,  Padhraic and Goulden,  Michael L. and Randerson,  James T.},
  year = {2022},
}

@article{balch_2020_fired,
  title = {{FIRED (Fire Events Delineation)}: {An} open, flexible algorithm and database of {US} fire events derived from the {MODIS} burned area product (2001–2019)},
  volume = {12},
  ISSN = {2072-4292},
  url = {https://doi.org/10.3390/rs12213498},
  number = {21},
  journal = {Remote Sensing},
  publisher = {MDPI AG},
  author = {Balch,  Jennifer K. and St. Denis,  Lise A. and Mahood,  Adam L. and Mietkiewicz,  Nathan P. and Williams,  Travis M. and McGlinchy,  Joe and Cook,  Maxwell C.},
  year = {2020},
  pages = {3498}
}

@article{andela_2019_atlas,
  title = {The {Global Fire Atlas} of individual fire size,  duration, speed and direction},
  volume = {11},
  ISSN = {1866-3516},
  url = {https://doi.org/10.5194/essd-11-529-2019},
  number = {2},
  journal = {Earth System Science Data},
  publisher = {Copernicus GmbH},
  author = {Andela,  Niels and Morton,  Douglas C. and Giglio,  Louis and Paugam,  Ronan and Chen,  Yang and Hantson,  Stijn and van der Werf,  Guido R. and Randerson,  James T.},
  year = {2019},
  pages = {529–552}
}

@misc{pyrostack_dataset,
  url = {https://doi.org/10.5281/zenodo.20434804},
  author = {Kondur,  Arya and Migliorini,  Giosue and Schmitt,  Cameron and Immorlano,  Francesco and Wang,  Tairan and Scholten,  Rebecca Christine and Foufoula-Georgiou,  Efi and Johnson,  Gary and Lautenberger,  Chris and Waeselynck,  Valentin and Romsos,  Jeffrey Shane and Shamsaei,  Kasra and Tejedor,  Alejandro and Liu,  Tianjia and Chen,  Yang and Smyth,  Padhraic and Randerson,  James},
  language = {en},
  title = {{PyroStack}},
  publisher = {Zenodo},
  year = {2026},
  copyright = {Creative Commons Attribution 4.0 International}
}

@article{schmit_2017_goesr,
  title = {A closer look at the {ABI} on the {GOES-R} series},
  volume = {98},
  ISSN = {1520-0477},
  url = {https://doi.org/10.1175/BAMS-D-15-00230.1},
  number = {4},
  journal = {Bulletin of the American Meteorological Society},
  publisher = {American Meteorological Society},
  author = {Schmit,  Timothy J. and Griffith,  Paul and Gunshor,  Mathew M. and Daniels,  Jaime M. and Goodman,  Steven J. and Lebair,  William J.},
  year = {2017},
  pages = {681–698}
}

@techreport{rothermel_1972_model,
  author       = {Rothermel, Richard C.},
  title        = {A mathematical model for predicting fire spread in wildland fuels},
  institution  = {USDA Forest Service, Intermountain Forest and Range Experiment Station},
  series       = {Research Paper},
  number       = {INT-115},
  address      = {Ogden, UT},
  year         = {1972},
  url = {https://doi.org/10.2737/INT-RP-115}
}

@article{TeymoorSeydi_2025_human_exposure,
  title = {Increasing global human exposure to wildland fires despite declining burned area},
  volume = {389},
  ISSN = {1095-9203},
  url = {https://doi.org/10.1126/science.adu6408},
  number = {6762},
  journal = {Science},
  publisher = {American Association for the Advancement of Science (AAAS)},
  author = {Teymoor Seydi,  Seyd and Abatzoglou,  John T. and Jones,  Matthew W. and Kolden,  Crystal A. and Filippelli,  Gabriel and Hurteau,  Matthew D. and AghaKouchak,  Amir and Luce,  Charles H. and Miao,  Chiyuan and Sadegh,  Mojtaba},
  year = {2025},
  pages = {826–829}
}

@article{liu_2024_progression,
  title = {Systematically tracking the hourly progression of large wildfires using {GOES} satellite observations},
  volume = {16},
  ISSN = {1866-3516},
  url = {https://doi.org/10.5194/essd-16-1395-2024},
  number = {3},
  journal = {Earth System Science Data},
  publisher = {Copernicus GmbH},
  author = {Liu,  Tianjia and Randerson,  James T. and Chen,  Yang and Morton,  Douglas C. and Wiggins,  Elizabeth B. and Smyth,  Padhraic and Foufoula-Georgiou,  Efi and Nadler,  Roy and Nevo,  Omer},
  year = {2024},
  pages = {1395–1424}
}

@inproceedings{kondylatos_2023_mesogeos,
 author = {Kondylatos, Spyridon and Prapas, Ioannis and Camps-Valls, Gustau and Papoutsis, Ioannis},
 booktitle = {{Advances in Neural Information Processing Systems}},
 editor = {A. Oh and T. Naumann and A. Globerson and K. Saenko and M. Hardt and S. Levine},
 pages = {50661--50676}, 
 title = {{Mesogeos}: {A} multi-purpose dataset for data-driven wildfire modeling in the {M}editerranean},
 year = {2023},
 url = {https://doi.org/10.52202/075280-2204},
 volume = {36},
}

@article{huot_2022_nextdaywildfirespread,
  author={Huot, Fantine and Hu, R. Lily and Goyal, Nita and Sankar, Tharun and Ihme, Matthias and Chen, Yi-Fan},
  journal={IEEE Transactions on Geoscience and Remote Sensing}, 
  title={{Next Day Wildfire Spread}: {A} machine learning dataset to predict wildfire spreading from remote-sensing data}, 
  year={2022},
  volume={60},
  pages={1-13},
  url={https://doi.org/10.1109/TGRS.2022.3192974}
}

@inproceedings{singla_2020_wildfiredb,
  title     = {{WildfireDB}: {A} Spatio-Temporal Dataset Combining Wildfire Occurrence with Relevant Covariates},
  author    = {Singla, Samriddhi and Diao, Tina and Mukhopadhyay, Ayan and Eldawy, Ahmed and Shachter, Ross and Kochenderfer, Mykel},
  booktitle = {{AI for Earth Sciences Workshop at NeurIPS 2020}},
  year      = {2020},
  pages     = {1--8},
  url       = {https://www.cs.ucr.edu/~eldawy/publications/20-ai4earth_neurips_WildfireDB.pdf}
}

@article{Li_2022_goes_frp,
  title = {Hourly biomass burning emissions product from blended geostationary and polar-orbiting satellites for air quality forecasting applications},
  volume = {281},
  ISSN = {0034-4257},
  url = {https://doi.org/10.1016/j.rse.2022.113237},
  journal = {Remote Sensing of Environment},
  publisher = {Elsevier BV},
  author = {Li,  Fangjun and Zhang,  Xiaoyang and Kondragunta,  Shobha and Lu,  Xiaoman and Csiszar,  Ivan and Schmidt,  Christopher C.},
  year = {2022},
  pages = {113237}
}

@article{Eidenshink2007_mtbs,
  title = {A Project for {Monitoring Trends in Burn Severity}},
  volume = {3},
  ISSN = {1933-9747},
  url = {https://doi.org/10.4996/fireecology.0301003},
  journal = {Fire Ecology},
  publisher = {Springer Science and Business Media LLC},
  author = {Eidenshink,  Jeff and Schwind,  Brian and Brewer,  Ken and Zhu,  Zhi-Liang and Quayle,  Brad and Howard,  Stephen},
  year = {2007},
  pages = {3–21}
}

@article{Picotte2020_mtbs_changes,
  title = {Changes to the {Monitoring Trends in Burn Severity} program mapping production procedures and data products},
  volume = {16},
  ISSN = {1933-9747},
  url = {https://doi.org/10.1186/s42408-020-00076-y},
  pages = {16},
  journal = {Fire Ecology},
  publisher = {Springer Science and Business Media LLC},
  author = {Picotte,  Joshua J. and Bhattarai,  Krishna and Howard,  Danny and Lecker,  Jennifer and Epting,  Justin and Quayle,  Brad and Benson,  Nate and Nelson,  Kurtis},
  year = {2020},
}

@misc{chen2026_fedsmtbs_dataset,
  url = {https://doi.org/10.5281/zenodo.20187963},
  author = {Chen,  Yang and Hantson,  Stijn and Andela,  Niels and Coffield,  Shane and Graff,  Casey and Morton,  Douglas and Ott,  Lesley and Foufoula-Georgiou,  Efi and Smyth,  Padhraic and Goulden,  Michael and Randerson,  James},
  title = {{FEDS-MTBS: An MTBS-constrained FEDS dataset in U.S.}},
  publisher = {Zenodo},
  year = {2026},
  copyright = {Creative Commons Attribution 4.0 International}
}

@article{rolph_2009_rave,
  title = {Description and verification of the {NOAA} smoke forecasting system: {The} 2007 fire season},
  volume = {24},
  ISSN = {0882-8156},
  url = {https://doi.org/10.1175/2008WAF2222165.1},
  number = {2},
  journal = {Weather and Forecasting},
  publisher = {American Meteorological Society},
  author = {Rolph,  Glenn D. and Draxler,  Roland R. and Stein,  Ariel F. and Taylor,  Albion and Ruminski,  Mark G. and Kondragunta,  Shobha and Zeng,  Jian and Huang,  Ho-Chun and Manikin,  Geoffrey and McQueen,  Jeffery T. and Davidson,  Paula M.},
  year = {2009},
  pages = {361–378}
}

@article{zhao_2025_tssatfire,
  title = {{TS-SatFire}: {A} multi-task satellite image time-series dataset for wildfire detection and prediction},
  volume = {12},
  ISSN = {2052-4463},
  url = {https://doi.org/10.1038/s41597-025-06271-3},
  pages = {1817},
  journal = {Scientific Data},
  publisher = {Springer Science and Business Media LLC},
  author = {Zhao,  Yu and Gerard,  Sebastian and Ban,  Yifang},
  year = {2025},
}

@inproceedings{lahrichi_2025_wsts+,
  title={Improved wildfire spread prediction with time-series data and the {WSTS+} benchmark},
  author={Lahrichi, Saad and Bova, Jake and Johnson, Jesse and Malof, Jordan},
  booktitle={{Proceedings of the IEEE/CVF Winter Conference on Applications of Computer Vision}},
  pages={2890--2900},
  year={2026},
  url = {https://doi.org/10.1109/WACV61042.2026.00283}
}

@misc{xu_2025_bcwildfire,
  title={{BCWildfire: A long-term multi-factor dataset and deep learning benchmark for boreal wildfire risk prediction}}, 
  author={Zhengsen Xu and Sibo Cheng and Lanying Wang and Hongjie He and Wentao Sun and Jonathan Li and Lincoln Linlin Xu},
  year={2025},
  eprint={2511.17597},
  archivePrefix={arXiv},
  primaryClass={cs.CV},
  url={https://arxiv.org/abs/2511.17597}, 
}

@article{barber_2024_canadianfirespread,
  title = {{The Canadian Fire Spread Dataset}},
  volume = {11},
  ISSN = {2052-4463},
  url = {https://doi.org/10.1038/s41597-024-03436-4},
  pages = {764},
  journal = {Scientific Data},
  publisher = {Springer Science and Business Media LLC},
  author = {Barber,  Quinn E. and Jain,  Piyush and Whitman,  Ellen and Thompson,  Dan K. and Guindon,  Luc and Parks,  Sean A. and Wang,  Xianli and Hethcoat,  Matthew G. and Parisien,  Marc-André},
  year = {2024},
}

@misc{bhowmik_2025_cawfi,
      title={{California Wildfire Inventory (CAWFI)}: {An} extensive dataset for predictive techniques based on artificial intelligence}, 
      author={Rohan Tan Bhowmik and Youn Soo Jung and Juan Aguilera and Mary Prunicki and Kari Nadeau},
      year={2025},
      eprint={2509.11015},
      archivePrefix={arXiv},
      primaryClass={cs.LG},
      url={https://arxiv.org/abs/2509.11015}, 
}

@article{lautenberger_2013_ELMFIRE,
    title = {Wildland fire modeling with an {Eulerian} level set method and automated calibration},
    journal = {Fire Safety Journal},
    volume = {62},
    pages = {289-298},
    year = {2013},
    issn = {0379-7112},
    url = {https://doi.org/10.1016/j.firesaf.2013.08.014},
    author = {Chris Lautenberger},
}

@inproceedings{ali_2024_dataset,
  author={Ali, Syed Haider and Zhang, Sunny and Mhatre, Saanvi and Xiao, Ting},
  booktitle={{2024 Conference on AI, Science, Engineering, and Technology (AIxSET)}}, 
  title={Advancing wildfire predictive models: {A} novel dataset for next-day wildfire spread}, 
  year={2024},
  pages={69-76},
  url={https://doi.org/10.1109/AIxSET62544.2024.00015}
}

@misc{erzibengoa_2025_iberfire,
      title={{IberFire -- A} detailed creation of a spatio-temporal dataset for wildfire risk assessment in {Spain}}, 
      author={Julen Erzibengoa and Meritxell Gómez-Omella and Izaro Goienetxea},
      year={2025},
      eprint={2505.00837},
      archivePrefix={arXiv},
      primaryClass={cs.LG},
      url={https://arxiv.org/abs/2505.00837}, 
}

@techreport{ncep2023rtma,
  author       = {{National Centers for Environmental Prediction}},
  title        = {Public Release Notes: {RTMA} v2.10.4 / {URMA} v2.10.3},
  institution  = {National Oceanic and Atmospheric Administration},
  year         = {2023},
  url          = {https://vlab.noaa.gov/documents/portlet_file_entry/715073/Public%2Brelease%2Bnotes%2BRTMA.v2.10_URMA.v2.10.pdf/090665f6-9815-6070-99d4-e140a25b7f18}
}

@article{depondeca_2011_noaa_analysis,
  title = {The {Real-Time Mesoscale Analysis} at {NOAA’s National Centers for Environmental Prediction}: {Current} status and development},
  volume = {26},
  ISSN = {1520-0434},
  url = {https://doi.org/10.1175/WAF-D-10-05037.1},
  number = {5},
  journal = {Weather and Forecasting},
  publisher = {American Meteorological Society},
  author = {De Pondeca,  Manuel S. F. V. and Manikin,  Geoffrey S. and DiMego,  Geoff and Benjamin,  Stanley G. and Parrish,  David F. and Purser,  R. James and Wu,  Wan-Shu and Horel,  John D. and Myrick,  David T. and Lin,  Ying and Aune,  Robert M. and Keyser,  Dennis and Colman,  Brad and Mann,  Greg and Vavra,  Jamie},
  year = {2011},
  pages = {593–612}
}

@article{jolly_2024_firedanger,
  title = {Modernizing the {US National Fire Danger Rating System} (version 4): {Simplified} fuel models and improved live and dead fuel moisture calculations},
  volume = {181},
  ISSN = {1364-8152},
  url = {https://doi.org/10.1016/j.envsoft.2024.106181},
  journal = {Environmental Modelling \& Software},
  publisher = {Elsevier BV},
  author = {Jolly,  W. Matt and Freeborn,  Patrick H. and Bradshaw,  Larry S. and Wallace,  Jon and Brittain,  Stuart},
  year = {2024},
  pages = {106181}
}

@techreport{rothermel_1991_rockymtfirs,
  title = {Predicting behavior and size of crown fires in the northern {Rocky Mountains}},
  url = {https://doi.org/10.2737/INT-RP-438},
  institution = {U.S. Department of Agriculture,  Forest Service,  Intermountain Research Station},
  author = {Rothermel,  Richard C.},
  year = {1991}
}

@article{schroeder_2014_FRP,
  title = {Integrated active fire retrievals and biomass burning emissions using complementary near-coincident ground,  airborne and spaceborne sensor data},
  volume = {140},
  ISSN = {0034-4257},
  url = {https://doi.org/10.1016/j.rse.2013.10.010},
  journal = {Remote Sensing of Environment},
  publisher = {Elsevier BV},
  author = {Schroeder,  Wilfrid and Ellicott,  Evan and Ichoku,  Charles and Ellison,  Luke and Dickinson,  Matthew B. and Ottmar,  Roger D. and Clements,  Craig and Hall,  Dianne and Ambrosia,  Vincent and Kremens,  Robert},
  year = {2014},
  pages = {719–730}
}

@misc{ELMFIREInOut,
  title   = {Basic inputs and outputs},
  author  = {Lautenberger, Chris},
  year    = {2025},
  url     = {https://elmfire.io/user_guide/io.html},
  urldate = {2025-06-02}
}

@misc{LANDFIREFuel,
  title   = {{LANDFIRE Fuel}},
  author  = {{United States Department of Agriculture}},
  year    = {2025},
  url     = {https://www.landfire.gov/fuel},
  urldate = {2025-06-02}
}

@misc{LANDFIREVeg,
  title   = {{LANDFIRE Vegetation}},
  author  = {{United States Department of Agriculture}},
  year    = {2025},
  url     = {https://www.landfire.gov/vegetation},
  urldate = {2025-06-02}
}

@misc{LANDFIRETopo,
  title   = {{LANDFIRE Topographic}},
  author  = {{United States Department of Agriculture}},
  year    = {2025},
  url     = {https://landfire.gov/topographic},
  urldate = {2025-06-02}
}

@misc{NPSFire_FireDanger,
  title   = {Understanding Fire Danger},
  author  = {{National Park Service, US Department of the Interior}},
  year    = {2025},
  url     = {https://www.nps.gov/articles/understanding-fire-danger.htm},
  urldate = {2025-06-02}
}

@misc{ECMWF_Parameters,
  title   = {Parameter Database},
  author  = {{European Centre for Medium-Range Weather Forecasts}},
  year    = {2025},
  url     = {https://codes.ecmwf.int/grib/param-db/},
  urldate = {2025-06-02}
}

@misc{cloudfire_worldgen_server,
  author       = {{CloudFire Inc.}},
  title        = {{CloudFire Worldgen Server}},
  year         = {2026},
  url = {https://worldgen.cloudfire.io},
}

@misc{PyretechnicsLibrary,
  title   = {The {Pyretechnics} Fire Behavior Library},
  author  = {{Spatial Informatics Group, LLC}},
  year    = {2026},
  url     = {https://pyregence.github.io/pyretechnics/#undefined},
  urldate = {2026-04-01}
}

@misc{pyrostack_github,
  url = {https://github.com/aryarksub/PyroStack},
  author = {Kondur,  Arya and Migliorini,  Giosue and Schmitt,  Cameron and Immorlano,  Francesco and Wang,  Tairan and Scholten,  Rebecca Christine and Foufoula-Georgiou,  Efi and Johnson,  Gary and Lautenberger,  Chris and Waeselynck,  Valentin and Romsos,  Jeffrey Shane and Shamsaei,  Kasra and Tejedor,  Alejandro and Liu,  Tianjia and Chen,  Yang and Smyth,  Padhraic and Randerson,  James},
  title = {{PyroStack}},
  year = {2026},
  publisher = {{GitHub}}
}

@article{keane_2005_cbd_estimate,
  title = {Estimating forest canopy bulk density using six indirect methods},
  volume = {35},
  ISSN = {1208-6037},
  url = {https://doi.org/10.1139/x04-213},
  number = {3},
  journal = {Canadian Journal of Forest Research},
  publisher = {Canadian Science Publishing},
  author = {Keane,  Robert E and Reinhardt,  Elizabeth D and Scott,  Joe and Gray,  Kathy and Reardon,  James},
  year = {2005},
  pages = {724–739}
}

@article{qi_2012_fuel_moisture,
  title = {Monitoring live fuel moisture using soil moisture and remote sensing proxies},
  volume = {8},
  ISSN = {1933-9747},
  url = {https://doi.org/10.4996/fireecology.0803071},
  number = {3},
  journal = {Fire Ecology},
  publisher = {Springer Science and Business Media LLC},
  author = {Qi,  Yi and Dennison,  Philip E. and Spencer,  Jessica and Riaño,  David},
  year = {2012},
  pages = {71–87}
}

@techreport{anderson_1982_fuelmodels,
  author = {Anderson,  Hal E.},
  title = {Aids to determining fuel models for estimating fire behavior},
  institution = {U.S. Department of Agriculture,  Forest Service,  Intermountain Forest and Range Experiment Station},
  number       = {INT-GTR-122},
  address      = {Ogden, UT},
  year         = {1982},
  url = {https://doi.org/10.2737/INT-GTR-122}
}

@Article{munoz-2021-era5land,
AUTHOR = {Mu\~noz-Sabater, J. and Dutra, E. and Agust\'{\i}-Panareda, A. and Albergel, C. and Arduini, G. and Balsamo, G. and Boussetta, S. and Choulga, M. and Harrigan, S. and Hersbach, H. and Martens, B. and Miralles, D. G. and Piles, M. and Rodr\'{\i}guez-Fern\'andez, N. J. and Zsoter, E. and Buontempo, C. and Th\'epaut, J.-N.},
TITLE = {{ERA5-Land}: {A} state-of-the-art global reanalysis dataset for land applications},
JOURNAL = {Earth System Science Data},
VOLUME = {13},
YEAR = {2021},
NUMBER = {9},
PAGES = {4349--4383},
URL = {https://doi.org/10.5194/essd-13-4349-2021}
}

@techreport{albini-baughman-1979,
AUTHOR = {Albini, Frank A. and Baughman, Robert G.},
TITLE = {Estimating windspeeds for predicting wildland fire behavior},
YEAR = {1979},
INSTITUTION = {U.S. Department of Agriculture, Forest Service, Intermountain Forest and Range Experiment Station},
TYPE        = {Research Paper},
NUMBER      = {INT-221},
ADDRESS     = {Ogden, UT},
url = {https://doi.org/10.5962/bhl.title.68710}
}

@techreport{scott-burgan,
AUTHOR = {Scott, Joe H. and Burgan, Robert E.},
TITLE = {Standard fire behavior fuel models: {A} comprehensive set for use with {Rothermel’s} surface fire spread model},
YEAR = {2005},
INSTITUTION = {U.S. Department of Agriculture, Forest Service, Rocky Mountain Research Station},
TYPE        = {General Technical Report},
NUMBER      = {RMRS-GTR-153},
ADDRESS     = {Fort Collins, CO},
url = {https://doi.org/10.2737/RMRS-GTR-153}
}

\newpage
\appendix

\section{Detailed PyroStack Layer Descriptions}\label{app:layer_desc}    

In this section, we provide a detailed description of each layer that can be found in the PyroStack dataset. As mentioned in the main text, layers are grouped into five categories based on data source, spatial resolution, and similar properties.

\subsection{Fire Characteristics}     

Fire characteristics layers capture information about how fires spread, including their burned area and fire radiative power. There are four such layers:
\begin{enumerate}
    \item \texttt{farea} (Fire area): This binary raster shows the estimated burned fire area at each FEDS update interval. A value of 1 means the cell lies within the burned area at that half-day time step; 0 means it does not. The area is derived via the FEDS clustering and alpha-hull method of VIIRS active fire detections every 12 hours \citep{chen_2022_viirs}.
    \item \texttt{fline} (Active fireline): This binary raster layer indicates where an active fire front was identified at a given 12-hour FEDS time step. A value of 1 means that this grid cell was flagged as part of the ``active fireline" (i.e., the portion of the fire perimeter adjacent to newly detected fire pixels) in that time step; a value of 0 means it was not. The underlying algorithm clusters VIIRS 375-m Active Fire Detections and estimates the fire object’s edge segments \citep{chen_2022_viirs}. The ``active fireline" corresponds to those burning parts of the perimeter at the time of the satellite overpass.
    \item \texttt{nfp} (New fire pixels): This binary raster layer indicates the cells where the newly detected fire pixels (from VIIRS) occurred in that FEDS time step \citep{chen_2022_viirs}. A cell with value 1 indicates that a new active-fire detection was recorded there during that interval; a value of 0 means no new fire pixel was detected in that cell.
    \item \texttt{frp} (Gridded fire radiative power): This layer quantifies the estimated fire radiative power (FRP) of the detected active fire pixels in each time step for each fire event tracked by FEDS. FRP is a satellite-derived measure of the rate of thermal energy being released by the fire at the time of the overpass \citep{schroeder_2014_FRP}. It essentially reflects how “intense” the burning is (higher FRP implies more combustion and higher heat release). The FRP for the most recent satellite overpass for each new fire pixel is reported for each grid cell, and in places where there are no new fire pixels, the data layer is assigned a value of 0.
\end{enumerate}

\subsection{Low-Resolution Climate}

Low-resolution climate layers capture dynamic information about temperature, rainfall, and pressure. There are four such layers:

\begin{enumerate}
    \item \texttt{d2m} (2-meter dewpoint temperature): This variable gives the dew point temperature (in Kelvin) of the air at a height of approximately 2 metres above the surface. In other words, it is the temperature to which air must be cooled at that height to become saturated (i.e., 100 \% relative humidity) given the current content of water vapor. It is derived by interpolating between the lowest model level and the surface \citep{ECMWF_Parameters}.
    \item \texttt{sp} (Surface pressure): This variable represents the total atmospheric pressure (in Pascals) exerted by the column of air directly above the surface of the Earth, measured at the model’s surface grid point. This is the weight of all the air in a column vertically above the area of the Earth’s surface represented at a fixed point \citep{ECMWF_Parameters}.
    \item \texttt{t2m} (2-meter temperature): This is the air temperature at  2 meters above the surface of land, sea, or in-land waters (in Kelvin). Temperature is calculated by performing interpolation between the lowest model level and the Earth's surface \citep{ECMWF_Parameters}.
    \item \texttt{tp} (Total precipitation): This variable corresponds to the accumulated total precipitation (liquid and solid forms) that has occurred over a given time interval to the surface. Precipitation covers both large-scale and convective (short-lived and intense) precipitation. Large-scale precipitation is generated by the cloud scheme in the Integrated Forecasting System (IFS). Convective precipitation is generated by the IFS convection scheme. Total precipitation does not include fog, dew, or evaporated precipitation. The data is reported in meters per hour which represent the depth of the precipitation if it were dispersed evenly over the entire grid box \citep{ECMWF_Parameters}.
\end{enumerate}

\subsection{High-Resolution Climate}

High-resolution climate layers capture dynamic (i.e. changing over time) information about fuel moisture content and wind. There are seven such layers:
\begin{enumerate}
    \item \texttt{lh} (Live herbaceous fuel moisture): This raster layer represents the moisture content of live herbaceous vegetation expressed as a percentage of the weight of that vegetation \citep{ELMFIREInOut}. For each grid cell, the value gives the water content relative to dry mass of the live fuel material. When herbaceous fuels are very green (high moisture values), they are less flammable. As moisture drops, fuel becomes more fire-prone \citep{NPSFire_FireDanger}. Excessively low moisture content indicates the vegetation can be treated as ``dead", whereas higher values correspond to ``live" vegetation \citep{qi_2012_fuel_moisture}.
    \item \texttt{lw} (Live woody fuel moisture): This raster layer quantifies the moisture content of live woody vegetation (such as shrubs, small trees, branches and foliage) given as a percentage of the weight of the material \citep{ELMFIREInOut}. Each cell gives the value for that location’s live woody fuel moisture. Low moisture content corresponds to dormant shrubs and increased dead fuel loads. On the other hand, high moisture content corresponds to fast-growing and expanding woody foliage.
    \item \texttt{m1} (1-hour dead fuel moisture): This raster layer provides the moisture content (in percent of weight) of the dead fine fuels with a 1-hour time-lag class \citep{ELMFIREInOut}. This typically corresponds to fuels with less than a quarter-inch diameter and the official classification is as "fine flashy fuels". For each cell, the value indicates how much water is in those very small dead fuel elements given current weather/fuel conditions. The "1-hour" time-lag means that it takes around an hour for these fuels to reach at least 63\% of equilibrium between its initial state and the current environment. The short time frame indicates quick responses to weather changes \citep{NPSFire_FireDanger}.
    \item \texttt{m10} (10-hour dead fuel moisture): This raster layer gives the moisture content (in percent of weight) of the dead fuel class with a $\sim$10-hour time-lag \citep{ELMFIREInOut}. This typically corresponds to rounded fuels or litter with a diameter in between a quarter-inch and an inch. Each cell’s value indicates the current moisture of these medium small dead fuels. These respond to weather changes more slowly than 1-hour fuels but more quickly than larger fuels \citep{NPSFire_FireDanger}.
    \item \texttt{m100} (100-hour dead fuel moisture): This raster describes the moisture content (in percent of weight) of dead fuels in the $\sim$100-hour timelag class \citep{ELMFIREInOut}. This typically corresponds to fields with a diameter ranging from 1 to 3 inches. The moisture content is calculated from 24-hour average boundary conditions composed of ecological and weather-based changes. Due to their larger size and thermal inertia, these fuels respond more slowly to changing environmental conditions \citep{NPSFire_FireDanger}.
    \item \texttt{wd} (20-foot wind direction): This raster layer represents the wind direction at the 20-foot height above the vegetative cover, expressed in degrees \citep{ELMFIREInOut}. Each cell gives the local wind direction at that height for that time step. Wind direction is reported as the direction from which the wind is blowing. In reality, wind direction can be influenced by local and general wind effects.
    \item {ws} (20-foot wind speed): This raster layer gives the wind speed in miles per hour at the standard height of 20-feet above the surface vegetation \citep{ELMFIREInOut}. For each grid cell, the value indicates the prevailing wind magnitude at that height and time step. The speed reported is the average over the provided area and may be influenced by both local and general wind effects.
\end{enumerate}

\subsection{Fuel Structure}

Fuel structure layers correspond to the static measurements of canopy features. There are four such layers:
\begin{enumerate}
    \item \texttt{cbd} (Canopy bulk density): This variable describes the mass of available canopy fuel per unit volume of canopy that would burn in a crown fire. A crown fire is an intense and/or fast-spreading fire that burns through the tops (canopies) of forests. Canopy bulk density essentially acts as a measure of how ``dense" the tree-canopy fuel is above the ground \citep{keane_2005_cbd_estimate}. In raster form, each cell gives an estimate of the canopy bulk density for that location, allowing spatial mapping of how canopy fuel structure changes across the landscape.
    \item \texttt{cbh} (Canopy base height): Canopy base height is the height above the ground at which the tree canopy fuel begins in a vegetation area (``stand"). In other words, it is the lowest height within the stand at which there is sufficient fuel continuity in the canopy layer to sustain fire propagation upward from the surface \citep{LANDFIREFuel}. Specifically, the data represents the average height from the ground to the stand’s canopy bottom \citep{ELMFIREInOut}. A higher CBH means a larger vertical gap between surface fuels and canopy fuels (i.e., less ladder fuel).
    \item \texttt{cc} (Canopy cover): Canopy cover (percent) refers to the proportion of ground surface area in each cell that is covered by the vertical projection of tree crowns \citep{LANDFIREFuel}. The raster form of this layer gives, for each cell, a value reflecting percent tree-crown coverage (e.g., 0-100 \%) \citep{ELMFIREInOut}. A dense canopy (high percent) implies more overhead fuel and possibly more intense crown fire behavior.
    \item \texttt{ch} {Canopy height}: This layer reflects the average height of the vegetative canopy (typically the uppermost live crown layer) for each cell \citep{ELMFIREInOut}. Taller canopies may support deeper flame structures, greater flame lengths in crown fire, and higher spotting potential \citep{LANDFIREFuel}. In raster form, each cell value gives the estimated canopy height.
\end{enumerate}

\subsection{Vegetation, Fuel Model, and Topography}

Layers taken from the LANDFIRE data source capture static vegetation, fuel model, and topographic environmental covariates. There are seven such layers:
\begin{enumerate}
    \item \texttt{*evt*} (Existing vegetation type): This raster layer represents the mapped distribution of current vegetation types across the landscape. Specifically, it uses the ecological‐systems classification developed by NatureServe for the Western hemisphere and groups of plant community types that occur under similar ecological conditions. EVT is also mapped using decision tree models, Landsat imagery, elevation, and biophysical gradient data. Each raster cell value contains a code referring to one of the hundreds of potential vegetation types \citep{LANDFIREVeg}.
    \item \texttt{*f[bfm]13*} (13 Anderson fire behavior fuel models): This layer is a raster classification of surface fuel model types based on the 13-model set defined by \cite{anderson_1982_fuelmodels} and adapted in LANDFIRE for modern conditions. The values in the raster correspond to one of the 13 fuel models and each model characterizes a distinct combination of dead and live fuel loadings, particle size distributions, fuel‐bed depth, moisture of extinction, etc. The fuel models are ultimately characterized by the most common fuel type for that particular combination of ecological characteristics \citep{LANDFIREFuel}.
    \item \texttt{*f[bfm]40*} (40 Scott and Burgan fire behavior fuel models): This raster layer encodes the fuel-model classification for surface fuels based on the Scott \& Burgan 40 Fire Behaviour Fuel Models (FBFM40). FBFM40 represents distributions of fuel loading found among different types of fuel components (live and dead), sizes, and types. Each cell’s integer value corresponds to one of the model codes that contain a fuel bed's parameters for behavior modeling. This fuel model set contains more classifications than related sets \citep{LANDFIREFuel}.
    \item \texttt{*roads*} (Operational roads): This raster layer represents the spatial locations of existing roads of various classes across the landscape. All pixels correspond to one of four road classes: primary roads (value 20), secondary roads (21), tertiary roads (22), and thinned roads (23). This layer is mostly used for supporting fire operations and infrastructure \citep{LANDFIREFuel}.
    \item \texttt{*asp*} (Topographic aspect): This raster records the aspect (direction) of the terrain slope for each grid cell, measured in degrees. Aspect defines the downslope direction in degrees and is derived from the Digital Elevation Model data. Aspect is also calculated from ``true north" (geographic North Pole) \citep{LANDFIRETopo}.
    \item \texttt{*elev*} (Topographic elevation): This layer represents the ground elevation in meters above sea level for each raster cell. Data is derived from the latest 1-arc‐second DEM data. Data is also calculated using ``true north" (geographic North Pole) \citep{LANDFIRETopo}.
    \item \texttt{*slpd*} (Topographic slope degrees): This raster layer gives the slope (steepness) of the terrain surface for each grid cell, measured in degrees. Essentially, this represents the change of elevation over a specific area. Larger values indicate more steepness, whereas values closer to zero indicate more flatness. Data is generated from 1-arc‐second digital elevation model tiles and each cell's slope is derived from the neighboring elevation values \citep{LANDFIRETopo}.
\end{enumerate}

\section{Vision Transformer Model Details}\label{app:vit_details}

To demonstrate the utility of PyroStack from a \gls{ml} perspective, we implement a Vision Transformer (ViT) model for one-step-ahead fire growth prediction. We adopt a ViT in encoder-only configuration with bidirectional self-attention. To improve generalization, we augment training with a masking strategy inspired by Masked Autoencoders (MAE) \citep{he2022masked} and masked diffusion models \citep{sahoo2024simple, shi2024simplified}: the target output is randomly masked with a special \texttt{[MASK]} token during training, concatenated with the conditioning inputs, and fed to the model. At inference, the output channel is fully masked, prompting the model to predict the complete area from conditioning information alone. 

\end{document}